\documentclass[10pt,twocolumn,letterpaper]{article}

\usepackage{caption}
\usepackage[subrefformat=parens]{subcaption}

\usepackage{footmisc}
\usepackage{makecell}

\usepackage{iccv}
\usepackage{times}
\usepackage{epsfig}
\usepackage{graphicx}
\usepackage{amsmath}
\usepackage{amssymb}
\usepackage{relsize}

\usepackage{booktabs}
\usepackage[inline]{enumitem}

\usepackage{siunitx}
\DeclareSIUnit\px{px}

\usepackage[super]{nth}

\usepackage{pifont}
\newcommand{\cmark}{\ding{51}}%
\newcommand{\xmark}{\ding{55}}%

\newcolumntype{H}{>{\setbox0=\hbox\bgroup}c<{\egroup}@{}}

\newcommand{\mAP}{\ensuremath{\mathrm{mAP}}}
\newcommand{\mAPs}{\ensuremath{\mathrm{mAP}_\mathrm{s}}}
\newcommand{\mAPm}{\ensuremath{\mathrm{mAP}_\mathrm{m}}}
\newcommand{\mAPl}{\ensuremath{\mathrm{mAP}_\mathrm{l}}}

\newcommand{\AP}{\ensuremath{\mathrm{AP}}}
\newcommand{\APs}{\ensuremath{\mathrm{AP}_\mathrm{s}}}
\newcommand{\APm}{\ensuremath{\mathrm{AP}_\mathrm{m}}}
\newcommand{\APl}{\ensuremath{\mathrm{AP}_\mathrm{l}}}

\newcommand{\mtsfull}{Mapillary Traffic Sign Dataset}
\newcommand{\mts}{MTSD}
\newcommand{\mpx}{MPixels}

\usepackage[pagebackref=true,breaklinks=true,letterpaper=true,colorlinks,bookmarks=false]{hyperref}
\usepackage[capitalise, nameinlink, noabbrev]{cleveref}

\usepackage{tikz}
\usetikzlibrary{spy}

\iccvfinalcopy 

\def\iccvPaperID{} 

\ificcvfinal\fi
\begin{document}

\title{The Mapillary Traffic Sign Dataset for Detection and Classification on a Global Scale}


\author{Christian Ertler \and Jerneja Mislej \and Tobias Ollmann \and Lorenzo Porzi \and Gerhard Neuhold \and Yubin Kuang\\
{\tt\small \emph{\{christian, jerneja, tobias, lorenzo, gerhard, yubin\}}@mapillary.com}
}

\twocolumn[{%
\renewcommand\twocolumn[1][]{#1}%
\maketitle
\begin{center}
    \centering
    \includegraphics[width=\linewidth]{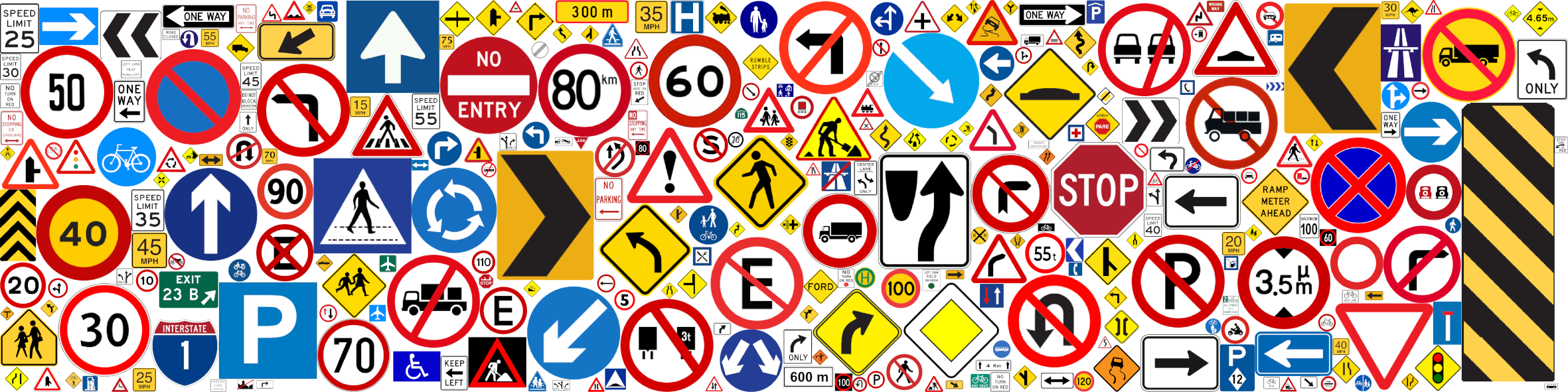}
    \captionof{figure}{Taxonomy overview. The sizes are relative to the number of samples within the Mapillary Traffic Sign Dataset (\mts).}
    \label{fig:catchy}
\end{center}%
}]

\begin{abstract}
Traffic signs are essential map features globally in the era of autonomous driving and smart cities. To develop accurate and robust algorithms for traffic sign detection and classification, a large-scale and diverse benchmark dataset is required. In this paper, we introduce a traffic sign benchmark dataset of \num{100}$K$ street-level images around the world that encapsulates diverse scenes, wide coverage of geographical locations, and varying weather and lighting conditions and covers more than \num{300} manually annotated traffic sign classes. The dataset includes \num{52}$K$ images that are fully annotated and \num{48}$K$ images that are partially annotated.  This is the largest and the most diverse traffic sign dataset consisting of images from all over world with fine-grained annotations of traffic sign classes. We have run extensive experiments to establish strong baselines for both the detection and the classification tasks. In addition, we have verified that the diversity of this dataset enables effective transfer learning for existing large-scale benchmark datasets on traffic sign detection and classification. The dataset is freely available for academic research\footnote{\url{www.mapillary.com/dataset/trafficsign}}.
\end{abstract}

\begin{table*}
\renewcommand*{\thefootnote}{\fnsymbol{footnote}}
\begin{tabular}{@{}lrrrclcc@{}}
\toprule
\textbf{Dataset} & \textbf{\#Images} & \textbf{\#Classes} & \textbf{\#Signs} & \textbf{Attributes} & \textbf{Region} & \textbf{BBoxes} & \textbf{Unique} \\ \midrule
\makecell[l]{\mts~(TRAIN+VAL) \\ \mts} & \makecell[r]{\num{41907} \\ \num{100000}} & 313 & \makecell[r]{\footnotemark[1]\num{206388} \\ \num{325172}} & \makecell[c]{occluded, exterior, \\ out-of-frame, dummy, \\ ambiguous, included} & \textbf{world-wide} & \cmark & \makecell[c]{\cmark \\ \xmark\footnotemark[4]} \\ \midrule
TT100K~\cite{zhu2016traffic} & \footnotemark[2] \num{100000} & \footnotemark[6] 221 & \num{26349}  & \xmark & China & \cmark & \cmark \\ 
MVD~\cite{neuhold2017mapillary} & \num{20000} & \footnotemark[2]2 & \num{174541} & \xmark & \textbf{world-wide} & \cmark & \cmark \\ \midrule
GTSDB~\cite{Houben-IJCNN-2013} & \num{900} & 43 &  \num{852} & \xmark & Germany & \cmark & \xmark \\
RTSD~\cite{rtsd} & \footnotemark[3]\num{179138}  & 156 & \footnotemark[3]\num{104358} & \xmark & Russia & \cmark & \xmark \\
STS~\cite{larssonSCIA2011} & \num{3777} & 20 & \num{5582} & \xmark & Sweden & \cmark & \xmark \\
LISA~\cite{mogelmose2012vision} & \num{6610} & 47 & \num{7855}  & \xmark & USA & \cmark & \xmark \\
GTSRB~\cite{Stallkamp2012} & \xmark & 43 & \num{39210} & \xmark & Germany & \xmark & \xmark \\
BelgiumTS~\cite{timofte2014multi} & \xmark & 108 & \num{8851} & \xmark & Belgium & \xmark & \xmark \\ \bottomrule
\end{tabular}
\caption{Overview of traffic sign datasets. The numbers include only publicly available images and annotations~(\eg we only report numbers for the training and validation set for \mts). \emph{Unique} refers to datasets where each traffic sign bounding box corresponds to a unique traffic sign instance (\ie no sequences showing the same sign again and again). \protect\footnotemark[1]\num{66138} signs are within the taxonomy. \protect\footnotemark[2] TT100K contains only \num{10000} images containing traffic signs. \protect\footnotemark[6] only 45 classes have more than 100 examples. \protect\footnotemark[2] MVD contains only back \emph{vs.}\ front classes. \protect\footnotemark[3]video-frames covering only \num{15630} unique signs. \protect\footnotemark[4]signs within the partially annotated set correspond to signs within the training set.}
\label{tab:dataset_comp}
\end{table*}

\section{Introduction}
Robust and accurate object detection and classification in diverse scenes is one of the essential tasks in computer vision. With the development and application of deep learning in computer vision, object detection and recognition has been studied~\cite{girshick2015fast, ren2015faster, lin2017focal} extensively on general scene understanding datasets~\cite{lin2014microsoft, everingham2015pascal, kuznetsova2018open}. In terms of fine-grained detection and classification, there are also the datasets that focused on general hierarchical object classes~\cite{kuznetsova2018open} or domain-specific datasets, \eg on bird species~\cite{wah2011caltech}. In this paper, we will focus on detection and fine-grained classification of traffic signs.

Traffic signs are key map features for navigation, traffic control, and road safety. Specifically, traffic signs encode information for driving directions, traffic regulation, and early warning. For autonomous driving, accurate and robust perception of traffic signs is essential for localization and motion planning. 

As an object class, traffic signs have specific characteristics in their appearance. First of all, traffic signs are in general rigid and planar. Secondly, traffic signs are designed to be distinctive from their surroundings. In addition, there is limited variety in colors and shapes for traffic signs. For instance, regulatory signs in European countries are typically circular with a red border. 
To some degree, the aforementioned characteristics limit the appearance variation and increase the distinctness of traffic signs. However, traffic sign detection and classification is still a very challenging problem due to the the following reasons: \begin{enumerate*}[label=(\arabic*)]
\item traffic signs are easily confused with other object classes in street scenes (\eg advertisements, banners, and billboards); 
\item reflection, low light condition, damages, and occlusion hinder the classification performance of a sign class; 
\item fine-grained classification with small inter-class difference is not trivial; 
\item the majority of traffic signs---when appearing in street-level images---are relatively small in size, which requires efficient architecture designs for small objects.
\end{enumerate*}

Traffic sign detection and classification has been studied extensively in the computer vision community. Specifically, convolutional neural networks (CNN)~\cite{sermanet2011traffic} have obtained great success for traffic sign classification in the German Traffic Sign Benchmark \cite{stallkamp2011german}.
Recent works on simultaneous detection and classification of traffic signs have also achieved good results on well-studied benchmark datasets~\cite{mathias2013traffic, zhu2016traffic} using either the Viola-Jones framework~\cite{viola2001rapid} or CNN-based methods~\cite{li2017perceptual}. However, these studies were done in relatively constrained settings in terms of the benchmark dataset:
the images and traffic signs are collected in a specific country; 
the number of traffic sign classes is relatively small;
the images lack diversity in weather conditions, camera sensors, and seasonal changes.

Extensive research is still needed for the task of detecting and classifying traffic signs at a global scale and under varying capture conditions and devices.  
In this paper, we present the following contributions: 
\begin{itemize} 
\item We present the most diverse traffic sign dataset with \num{100}$K$ images from all over the world. The dataset contains over \num{52}$K$ images that are fully annotated, covering \num{313} known traffic sign classes and other unknown classes, resulting in over \num{250}$K$ signs in total. Additionally, the dataset also includes about 48K images, where traffic signs are partially annotated by automatically propagating labels between neighboring images. 
\item We establish extensive baselines for detection and classification on the dataset, shedding light on future research directions.  
\item We study the impact of transfer learning using our traffic sign dataset and other traffic sign datasets released in the past.
Our results show that pre-training on our dataset boosts the average precision (\AP) of the binary detection task by \SIrange{4}{6}{\percent}, thanks to the completeness and diversity of our dataset. 
\end{itemize}

\noindent \textbf{Related Work.}
Traffic sign detection and recognition has been studied extensively in the literature in the past. The German Traffic Sign Benchmark Dataset (GTSBD)~\cite{stallkamp2011german} is one of the first datasets that was created to evaluate the classification branch of the problem. Following that, there have also been other traffic sign datasets focusing on regional traffic signs, \eg Swedish Traffic Sign Dataset~\cite{larsson2011using}, Belgium Traffic Sign Dataset~\cite{mathias2013traffic}, Russian Traffic Sign Dataset~\cite{shakhuro2016russian}, and Tsinghua-Tencent Dataset~(TT100K) in China~\cite{zhu2016traffic}.
For generic traffic sign detection (where no class information of the traffic signs is available), there has been work done in the Mapillary Vistas Dataset (MVD)~\cite{neuhold2017mapillary} (global) and BDD100K~\cite{yu2018bdd100k} (US only). A detailed overview and comparison of publicly available traffic sign datasets can be found in \cref{tab:dataset_comp}.


For general object detection, there has been substantial work on CNN-based methods with two main directions, \ie one-stage detectors~\cite{liu2016ssd, redmon2017yolo9000, lin2017focal} and two-stage detectors~\cite{girshick2014rich, girshick2015fast, ren2015faster, dai2016r}. One-stage detectors are generally much faster, trading off accuracy compared to two-stage detectors.
One exception is the one-stage RetinaNet~\cite{lin2017focal} architecture that outperforms the two-stage Faster-RCNN~\cite{ren2015faster} thanks to a weighting scheme during training to suppress trivial negative supervision.
For simultaneous detection and classification, one of the recent works~\cite{cheng2018revisiting} shows that decoupling the classification from detection head boosts the accuracy significantly.
Our work is related to~\cite{cheng2018revisiting} as we also decouple the detector from the traffic sign classifier.

To handle the scale variation of objects in the scene, many efficient multi-scale training and inference algorithms have been proposed and evaluated on existing datasets. For multi-scale training, in~\cite{singh2018analysis, singh2018sniper, li2019scale}, a few schemes have been proposed to distill supervision information from different scales efficiently by selective gradient propagation and crop generation. To enable efficient multi-scale inference, feature pyramid networks (FPN)~\cite{lin2017feature} were proposed to utilize lateral connections in a top-down architecture to construct effective multi-scale feature pyramid from a single image. 


To develop the baselines presented in this paper, we have chosen Faster-RCNN~\cite{ren2015faster} with FPN~\cite{lin2017feature} as the backbone. Given the aforementioned characteristics of traffic sign imagery, we have also trained a separate classifier for fine-grained classification as in \cite{cheng2018revisiting}. We elaborate the details of our baseline method in \cref{sec:detection}.

\section{\mtsfull}

In this section, we present a large-scale traffic sign dataset called \mtsfull ~(\mts) including \num{52453} images with fully annotated traffic sign bounding boxes and corresponding class labels. Additionally, it includes a set of \num{47547} nearby images with \num{87358} automatically generated labels, making it a total of \num{100000} images.
In the following sections we describe how the dataset was created and present our traffic sign class taxonomy consisting of 313 classes.

\subsection{Image Selection}
\label{sec:image_selection}

\begin{figure}
    \centering
    \includegraphics[width=\columnwidth]{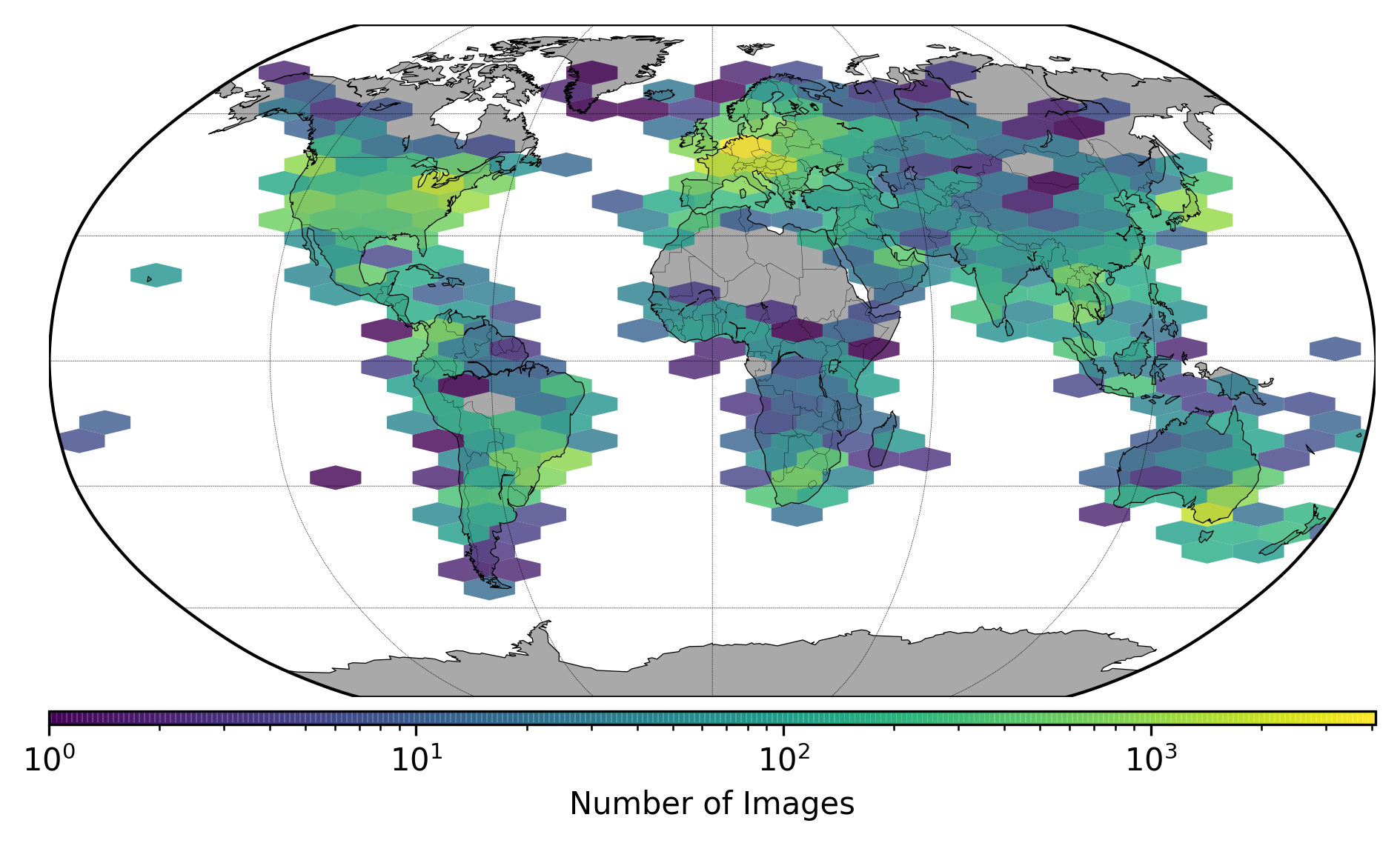}
    \caption{Geographical distribution of the images within \mts.}
    \label{fig:geo_distribution}
\end{figure}

There are various conventions for traffic signs in different parts of the world, leading to strong appearance differences.
Even within a single country or state, the distribution of signs is not uniform: some signs occur only in urban areas, some only on highways, and others only in rural areas.
With \mts, we present a dataset that covers this diversity uniformly. 
In order to do so, a proper pre-selection of images for annotation is crucial.
The requirements for this selection step are:
\begin{enumerate*}[label=(\arabic*)]
    \item to have a uniform geographical distribution of images around the world,
    \item to cover images of different quality, captured under varying conditions,
    \item to include as many signs as possible per image, and
    \item to compensate for the long-tailed distribution of potential traffic sign classes.
\end{enumerate*}

\emph{Mapillary}\footnote{\url{www.mapillary.com/app}} is a street-level imagery platform hosting images collected by members of their community. All images are accessible to everyone via Mapillary's public API under a CC-BY-SA license. Derived data including traffic sign detections are also available for academic research and approved applications. 

In order to get a pool of pre-selected images satisfying the aforementioned requirements, we sample images in a per-country manner with a greedy scheme.
The fraction of target images for each country is derived from the number of images available in that country and its population count weighted by a global target distribution over all continents (\ie \SI{20}{\percent} North America, \SI{20}{\percent} Europe, \SI{20}{\percent} Asia, \SI{15}{\percent} South America, \SI{15}{\percent} Oceania, \SI{10}{\percent} Africa).
We further make sure to cover both rural and urban areas within each country by binning the sampled images uniformly in terms of their geographical locations and sample random images from each of the resulting bins.
In the last step of our greedy image sampling scheme, we prioritize images containing at least one traffic sign instance according to the traffic sign detections given by the {Mapillary} API and make sure to cover various image resolutions, camera manufacturers and scene properties\footnote{Details on how scene properties are defined and derived are included in the supplementary materials.}. 
Additionally, we add a distance constraint so that selected images are far away from each other in order to avoid highly correlated images and traffic sign instances. 

The heat map in \cref{fig:geo_distribution} shows the resulting geographical distribution of the images within the dataset, covering almost all habitable areas of the world with higher density in populous areas.
Statistics of the final dataset can be found in \cref{sec:statistics}.

\subsection{Annotation Process}
\label{sec:annot_process}

The process of annotating an image including image selection approval, traffic sign localization by drawing bounding boxes, and class label assignment for each box is a complex and demanding task. To improve efficiency and quality, we split it into 3 consecutive tasks, with each having its own quality assurance process.
All tasks were done by \num{15} experts in image annotation after being trained with explicit specifications for each task.

\begin{description}[style=unboxed,leftmargin=0cm,listparindent=\parindent]
    \item[Image Approval.]
    Since initial image selection was done automatically based on the greedy heuristics described in \cref{sec:image_selection}, the annotators needed to reject images that did not fulfill our criteria for the dataset.
    In particular, we do not include non-street level images or images that have been taken from unusual places or viewpoints.
    Also we discarded images of very bad quality that could not be used for training at all (\ie extremely blurry or overexposed). 
    However, we still sample images of low quality in the dataset which include recognizable traffic signs as these are good examples to evaluate recognition on traffic signs in real-world scenarios.
    
    \item[Sign Localization.]
    In this task, the annotators were instructed to localize all traffic signs in the images and annotate them with bounding boxes.
    In contrast to previous traffic sign datasets where only specific types of traffic signs have been annotated (\eg TT100K~\cite{zhu2016traffic} includes only standard circular and triangular shaped signs), \mts\ contains bounding boxes for all types of traffic related signs including direction, information, highway signs, \etc.
    
    To speed up the annotation process, each image was initialized with bounding boxes of traffic signs extracted from the \emph{Mapillary} API.
    The annotators were asked to correct all existing bounding boxes to tightly contain the signs (or reject them in cases of false positives) and to annotate all missing traffic signs if their shorter sides were larger than \SI{10}{pixels}.
    We provide a statistical evaluation of the manual changes done by the annotators in \cref{sec:human_changes}.

    \item[Sign Classification.]
    This task was done independently for each annotated traffic sign.
    Each traffic sign (together with some image context) was shown to the annotators who were asked to provide the correct class label.
    This is not trivial, since the number of possible traffic sign classes is large. To the best of our knowledge, there is no globally valid traffic sign taxonomy available; even then, it would be impossible for the annotators to keep track of all the different traffic sign classes.
    
    To overcome this issue, we used a set of previously harvested template images of traffic signs from Wikimedia Commons~\cite{wikimedia} and grouped them by similarity in appearance and semantics.
    This set of templates (together with their grouping) defines the possible set of traffic sign classes that can be selected by the annotators.
    In fact, we store an identifier of the actual selected template, which allows us to link the traffic sign instances to our flexible traffic sign taxonomy without even knowing the final set of classes beforehand (see \cref{sec:class_taxonomy}).
    
    Since it would still be too time-consuming to scroll through the entire list of templates to choose the correct one out of thousands, we trained a neural network (with the grouped template images) to predict the similarities between an arbitrary image of a traffic sign instance and the templates.
    We used this proposal network to assist the annotators in choosing the correct template by pre-sorting the template list for each individual traffic sign.
    For cases in which this strategy fails to provide a matching template, we provided a text-based search for templates.
    For details about the annotation UI and how the proposal network was used to assist the annotator we refer to the supplemental material.
    
    \item[Additional Attributes.]
    In addition to the bounding boxes and the matching traffic sign templates, the annotators were asked to provide additional attributes for each sign: \emph{occluded} if the sign is partly occluded; \emph{ambiguous} if the sign is not classifiable at all (\eg too small, bad quality, heavy occlusion \etc);
    \emph{dummy} if it looks like a sign but isn't (\eg car stickers, reflections, \etc); \emph{out-of-frame} if the sign is cut off by the image border; \emph{included} if the sign is part of another bigger sign; and \emph{exterior} if the sign includes other signs.
    Some of these attributes were assigned during localization (if context information is needed). The rest was assigned during classification.
    In \cref{sec:detection} we describe  how we use some of these attributes to guide the training of our traffic sign detector. 
    
    \item[Annotation Quality.]
    All annotations in \mts\ were done by expert annotators going through a thorough training process. Their work was monitored by a continuous quality control (QC) process to quickly identify problems during annotations. Moreover, our step-wise annotation process (\ie approval followed by localization followed by classification) ensures that each traffic sign was seen by at least two annotators. The second annotator operating in the classification step was able to reject false positive signs or to report issues with the bounding box in which case the containing image was sent back to the localization step.
    
    In additional quality assurance (QA) experiments done by a \nth{2} annotator on \num{5}$K$ images including \num{26}$K$ traffic signs we found that
    \begin{itemize}
        \item only \SI{0.5}{\percent} of bounding boxes needed correction.
        \item the false negative rate was \SI{0.89}{\percent} (corresponding to a total number of only \num{212} missing signs; most of them being very small).
        \item the false positive rate was at \SI{2.45}{\percent}. Note that is in the localization step before classification, where a second annotator would have been asked to classify the sign and could potentially fix false positives.
    \end{itemize}
    
\end{description}

\subsection{Traffic Sign Class Taxonomy}
\label{sec:class_taxonomy}

Traffic signs vary across different countries. For many countries, there exists no publicly available and complete catalogue of signs.
The lack of a known set of traffic sign classes leads to challenges in assigning class labels to traffic signs annotated in \mts. 
The potential magnitude of this unknown set of traffic signs is in the thousands as indicated by the set of template images described in \cref{sec:annot_process}.

\begin{figure}
    \centering
    \includegraphics[width=0.85\linewidth]{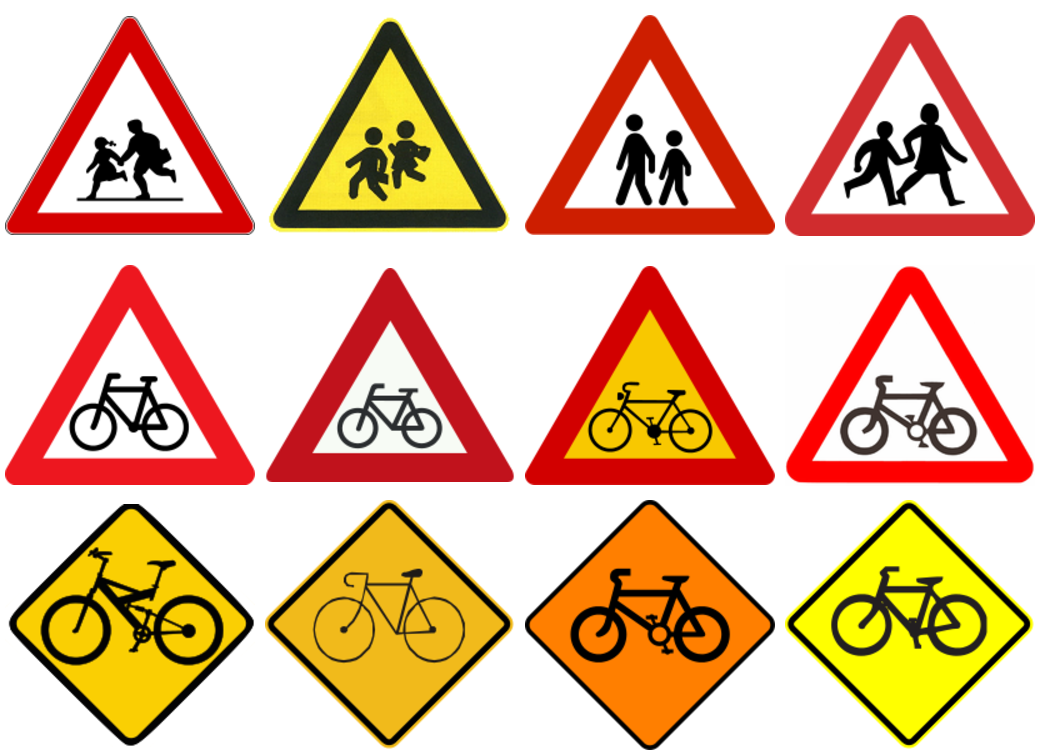}
    \caption{Example templates in traffic sign taxonomy. Each row represent a traffic sign class based on semantics and appearance.}
    \label{fig:vis_taxonomy}
\end{figure}

For \mts, we did a manual inspection of the templates that have been chosen by the annotators and selected a subset of them to form the final set of \num{313} classes included in the dataset as visualized in \cref{fig:catchy}.
This subset was chosen and grouped such that there are no overlaps or confusion (visual or semantic) among the classes.
Templates with the same semantics and similar appearance form a class. However, different groups of templates that share the same semantics but are different in terms of appearance form different classes as shown in ~\cref{fig:vis_taxonomy}.

All these classes defined by disjoint sets of templates build up our traffic sign class taxonomy.
We map all annotated traffic signs in \mts\ that have a template selected within this taxonomy to a class label.
We would like to emphasize that our flexible traffic sign taxonomy allows us to incrementally extend \mts\ by adding more classes together with already annotated traffic sign instances with known templates.

\subsection{Partial Annotations}
\label{sec:partial}

In addition to the fully-annotated images, we provide another set of images with partially annotated bounding boxes and labels of traffic signs.
Given the fully-annotated images, the annotations of this set of images are generated automatically. We achieve this by finding correspondences between the manual annotations in the fully-annotated images and automatic detections in geographically neighboring images from the {Mapillary} API. To find these correspondences, we first use Structure from Motion (SfM)~\cite{hartley2003multiple} to recover the relative camera poses between the fully-annotated images and the partially annotated images. With these estimated relative poses, we generate the correspondences between annotated signs and automatically detected signs by triangulating and verifying the re-projection errors for the centers of the bounding boxes between multiple images. Having these correspondences, we propagate the manually annotated class labels to the automatic detections in the partially annotated images. Since there is no guarantee that all traffic signs are detected through {Mapillary}'s platform, we obtained a set of images with partially annotated bounding box annotations and labels. Note that, for unbiased evaluation, we ensure that the extension is done only in the geographical neighborhood of images in the training set (based on the split discussed in \cref{sec:data_split}). Example images can be found in \cref{fig:partial_images}.
A more detailed description of how this set was created can be found in supplemental material.


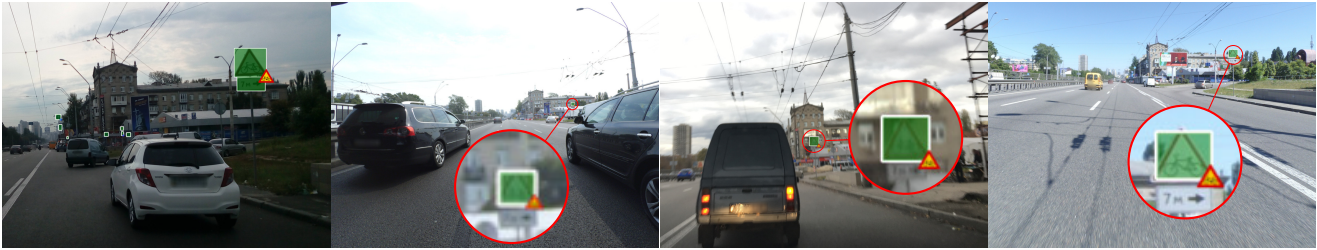
\begin{figure*}
    \centering
    \hfill\begin{subfigure}[t]{0.249\linewidth}
        \centering
        \begin{tikzpicture}[spy using outlines={circle,red,magnification=10,size=1.5cm, connect spies}]
            \node[anchor=south west,inner sep=0] (image){\includegraphics[width=\linewidth]{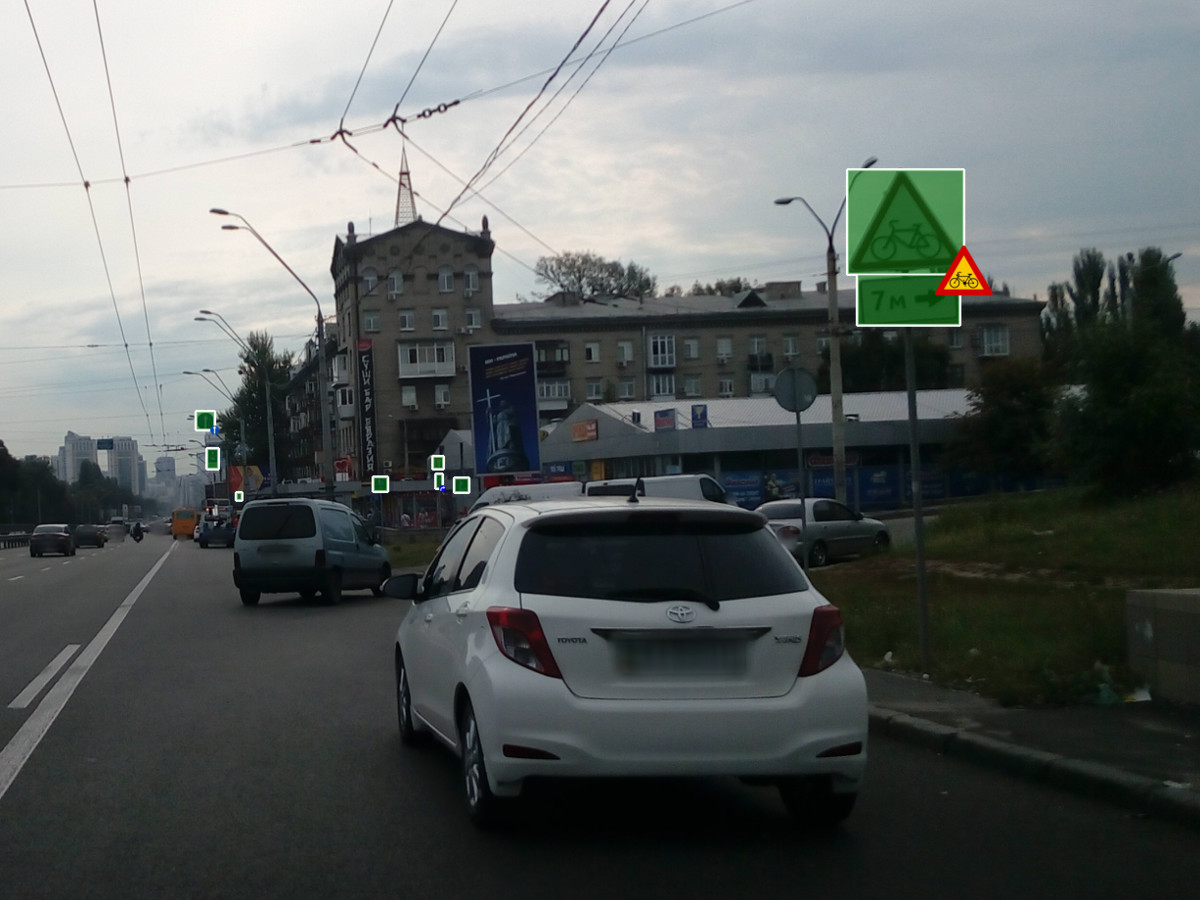}};
            \begin{scope}[x={(image.south east)},y={(image.north west)}]
            \end{scope}
        \end{tikzpicture}
    \end{subfigure}\hfill%
    \begin{subfigure}[t]{0.249\linewidth}
        \centering
        \begin{tikzpicture}[spy using outlines={circle,red,magnification=10,size=1.5cm, connect spies}]
            \node[anchor=south west,inner sep=0] (image){\includegraphics[width=\linewidth]{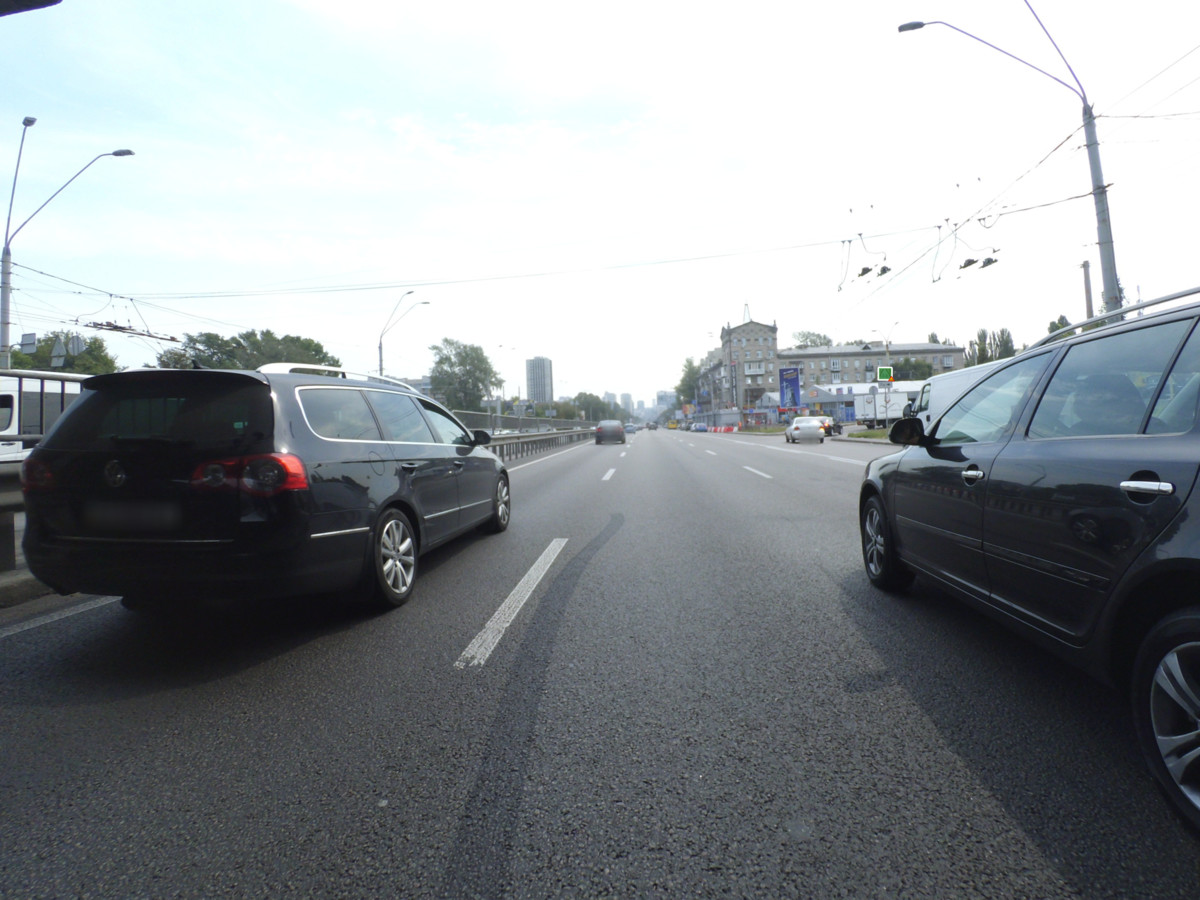}};
            \begin{scope}[x={(image.south east)},y={(image.north west)}]
                \coordinate (pos spy) at (.55,.25);
                \coordinate (center) at (.735,.585);
                \spy on (center) in node [] at (pos spy);
            \end{scope}
        \end{tikzpicture}
    \end{subfigure}\hfill%
    \begin{subfigure}[t]{0.249\linewidth}
        \centering
        \begin{tikzpicture}[spy using outlines={circle,red,magnification=5,size=1.5cm, connect spies}]
            \node[anchor=south west,inner sep=0] (image){\includegraphics[width=\linewidth]{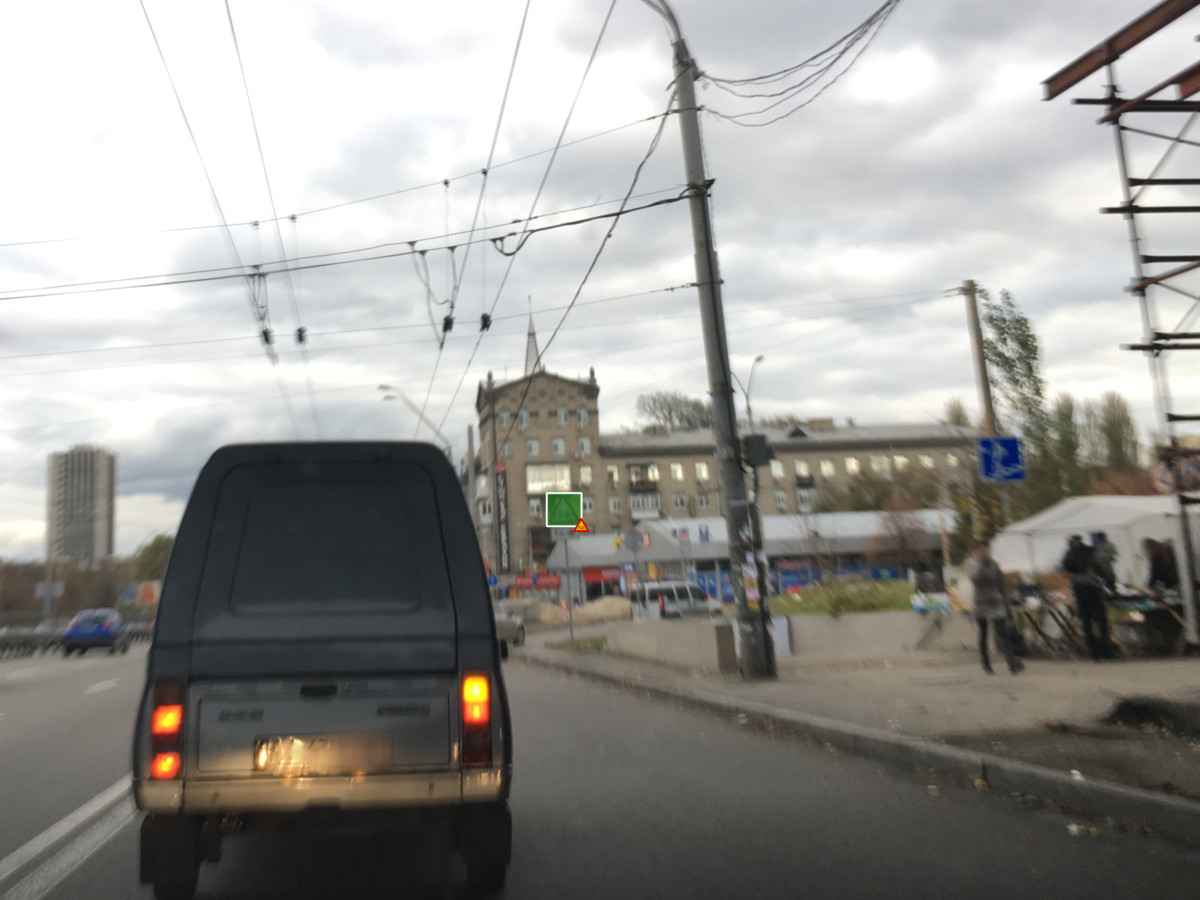}};
            \begin{scope}[x={(image.south east)},y={(image.north west)}]
                \coordinate (pos spy) at (.75,.45);
                \coordinate (center) at (.47,.435);
                \spy on (center) in node [] at (pos spy);
            \end{scope}
        \end{tikzpicture}
    \end{subfigure}\hfill%
    \begin{subfigure}[t]{0.249\linewidth}
        \centering
        \begin{tikzpicture}[spy using outlines={circle,red,magnification=6,size=1.5cm, connect spies}]
            \node[anchor=south west,inner sep=0] (image){\includegraphics[width=\linewidth]{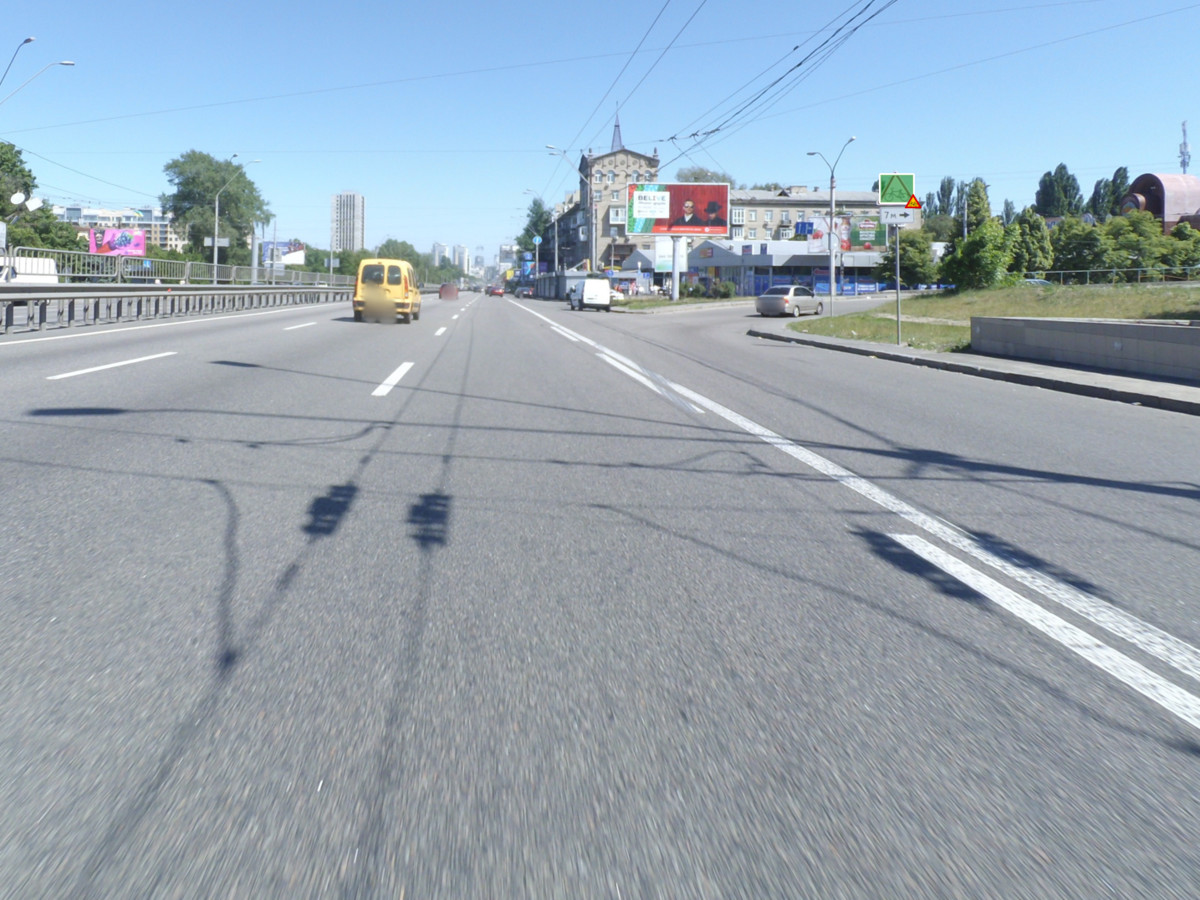}};
            \begin{scope}[x={(image.south east)},y={(image.north west)}]
                \coordinate (pos spy) at (.6,.35);
                \coordinate (center) at (.747,.785);
                \spy on (center) in node [] at (pos spy);
            \end{scope}
        \end{tikzpicture}
    \end{subfigure}\hfill%

    \caption{Example from the partially annotated set: The leftmost image is from the fully annotated set. The other 3 images show the same sign from different perspectives in the partially annotated set. Best viewed zoomed in.}  
    \label{fig:partial_images}
\end{figure*}

We see this set of partially annotated images serving as a data source for further research in semi-supervised learning for traffic sign detection and recognition. In addition, the correspondence information between the traffic sign observations will pave the ways for other learning tasks like semantic matching with street-level objects.

\subsection{Dataset Splits}
\label{sec:data_split}

As common practice with other datasets such as COCO~\cite{lin2014microsoft}, MVD~\cite{neuhold2017mapillary} and PASCAL~VOC~\cite{everingham2015pascal}, we split \mts\ into training, validation and test sets, consisting of \num{36589}, \num{5320}, and \num{10544} images, respectively.
We provide the image data for all images as well as the annotations for the training and validation set; the annotations for the test set will not be released in order to ensure a fair evaluation.
Additionally, we provide a set of \num{47547} images with partial annotations as discussed in \cref{sec:partial} that can be used for training as well.

Each split is created in a way to match the distributions described in \cref{sec:image_selection}.
Especially, we ensure that the distribution of class instances is similar for each split, to avoid that rare classes are under-represented in the smaller sets (\ie. validation/test sets).
The same holds true for the additional sign attributes (\eg \emph{ambiguous}, \etc).

\section{Statistics}
\label{sec:statistics}

In this section, we provide statistics of image and traffic sign properties of \mts\ and also a comparison with previous datasets such as TT100K~\cite{zhu2016traffic} and MVD~\cite{neuhold2017mapillary}.
Furthermore, we provide statistical insights about annotator actions during the creation of \mts.

\subsection{Image Properties}

\begin{figure}
    \centering
    \includegraphics[width=0.9\linewidth]{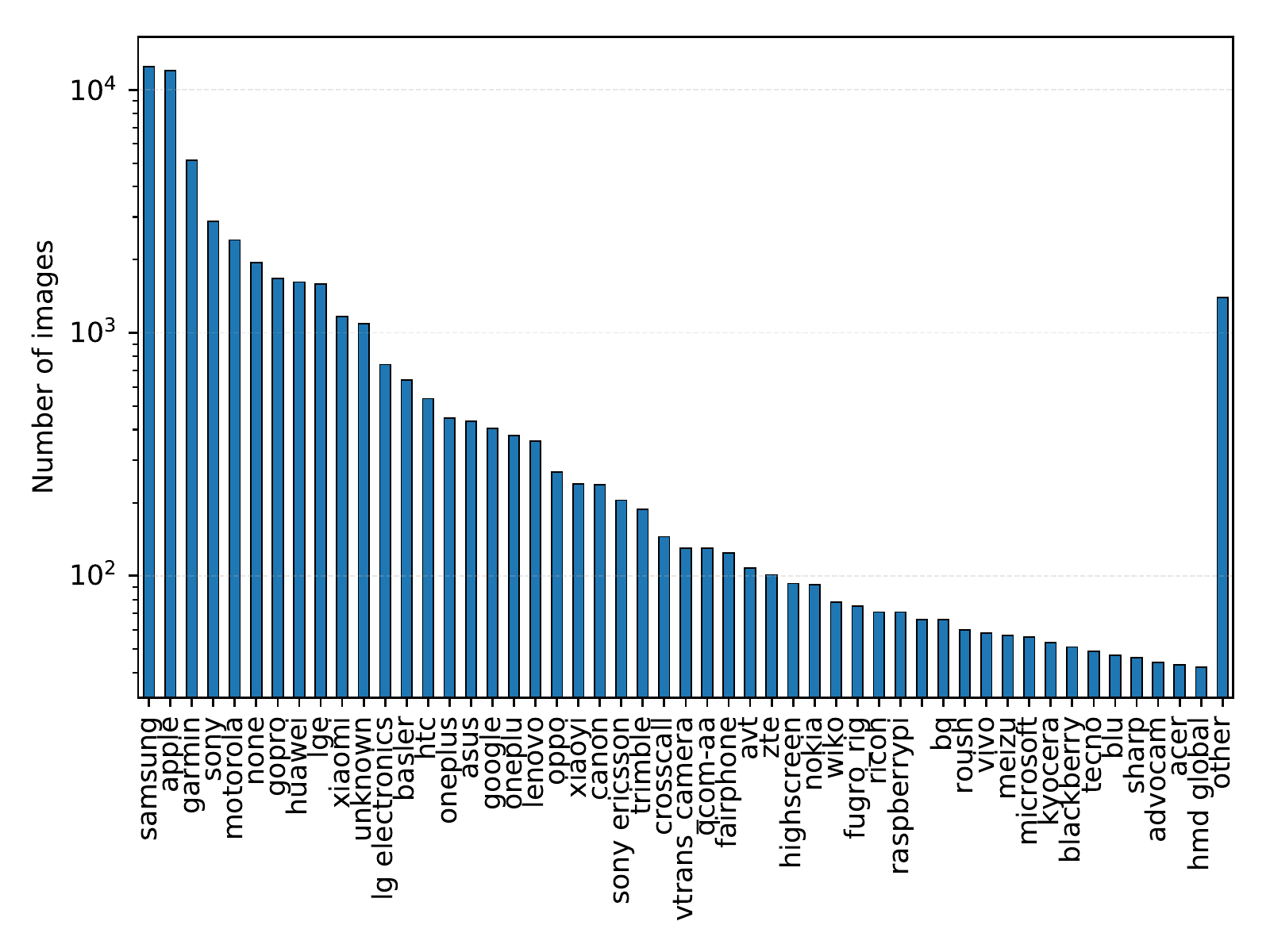}
    \caption{Distribution of camera devices used for image capture.}
    \label{fig:make_distribution}
\end{figure}

For a diverse dataset to reflect a real-world image capturing setting, it is important to cover a broad range of different image qualities and other image properties such as aspect ratio, focal length and other sensor-specific properties.
The image selection strategy described in \cref{sec:image_selection} used for \mts\ ensures a good distribution over different capturing settings.
In \cref{fig:make_distribution}, we show the distribution of camera manufacturers used for capturing the images of \mts.
In total, the dataset covers over \num{200} different sensor manufacturers (we group the tail of the distribution for displaying purposes) which results in a large variety of image properties similar to the properties described in~\cite{neuhold2017mapillary}.
This is in contrast to the setup used for the TT100K~\cite{zhu2016traffic} which contains only images taken by a single sensor setup, making \mts\ more challenging in comparison.

\begin{figure}
    \centering
    \includegraphics[width=0.9\linewidth]{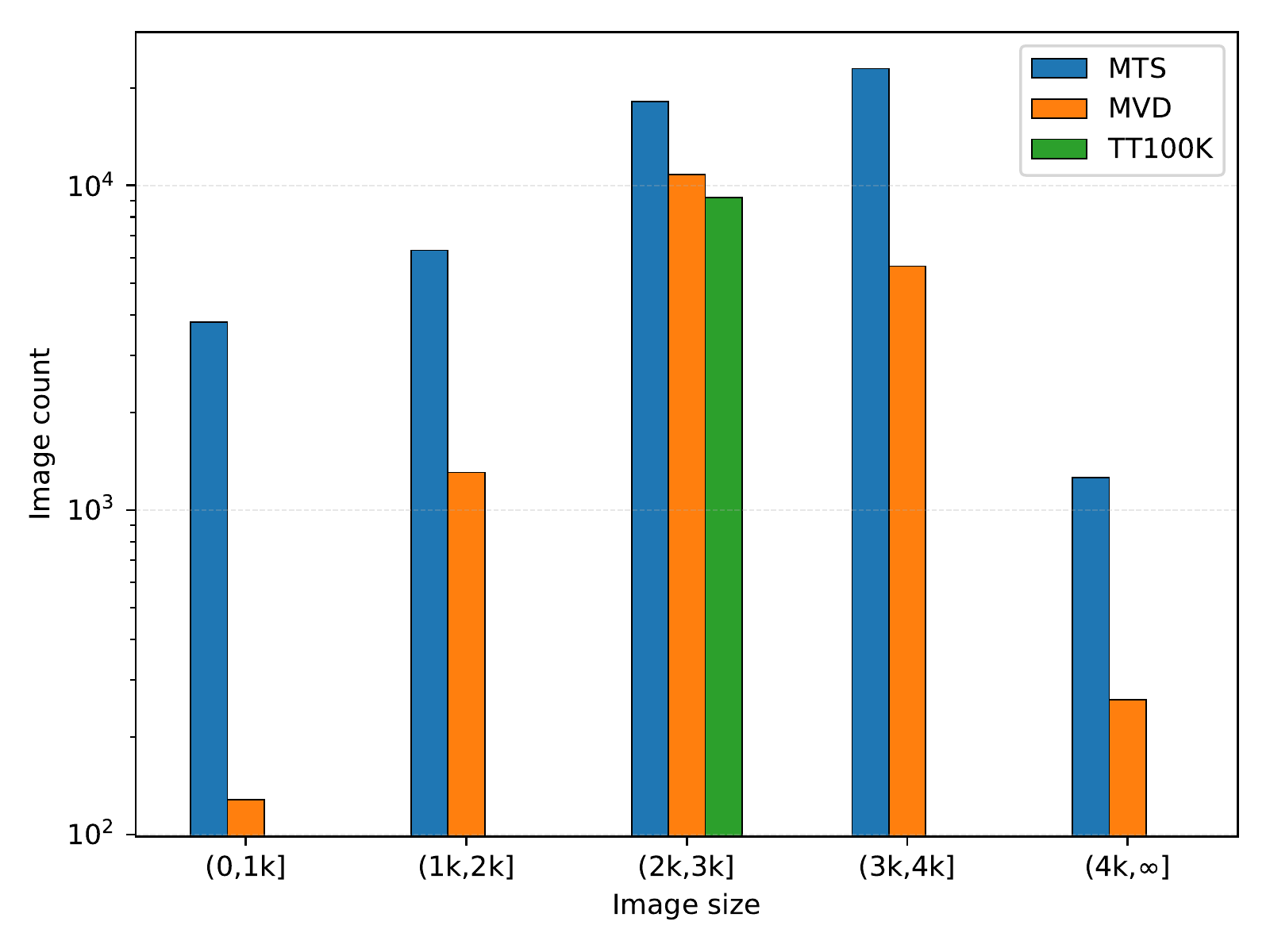}
    \caption{Distribution of image sizes (square root of pixel area).}
    \label{fig:image_sizes}
\end{figure}

The diversity in camera sensors further results in a diverse distribution over image resolutions as shown in \cref{fig:image_sizes}.
\mts\ covers a broad range of image sizes starting from low-resolution images with \SI{1}{\mpx\ } going up to images of more than \SI{16}{\mpx}.
Additionally, we include \num{1138} 360-degree panoramas stored as standard images with equi-rectangular projection.
Besides the overall larger image volume compared to other datasets, \mts\ also covers a larger fraction of low-resolution images, which is especially interesting for pre-training and validating detectors applied on similar sensors e.g. built-in automotive cameras.
For comparison, TT100K only contains images of \SI[parse-numbers=false]{2048^2}{\px} and even for this resolution the volume of images is smaller than in \mts.

\subsection{Traffic Sign Properties}
\label{sec:stat_compare}

\begin{figure*}
    \centering
    \begin{subfigure}[t]{0.33\linewidth}
        \includegraphics[width=\linewidth]{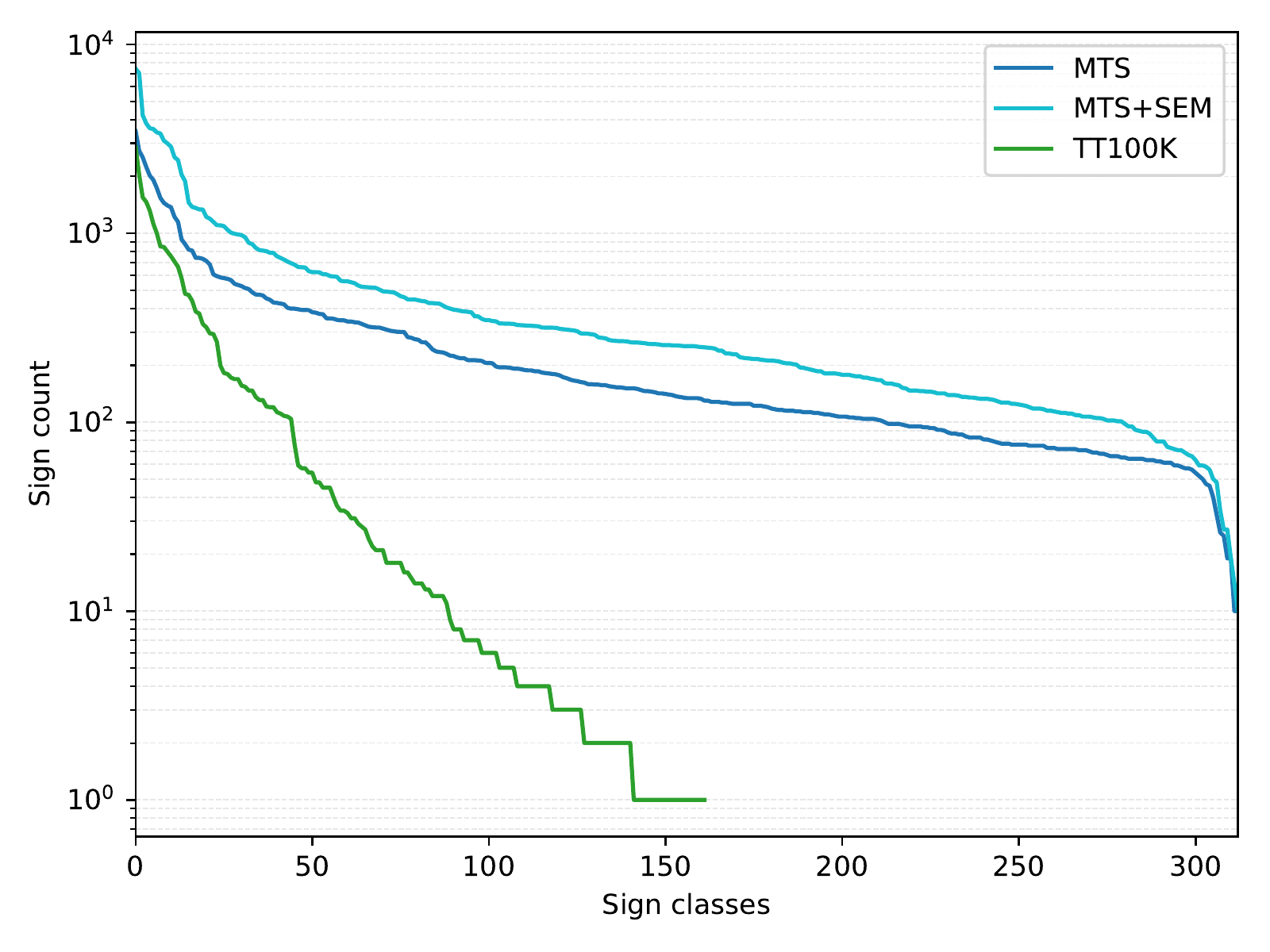}
    \end{subfigure}\hfill%
    \begin{subfigure}[t]{0.33\linewidth}
        \includegraphics[width=\linewidth]{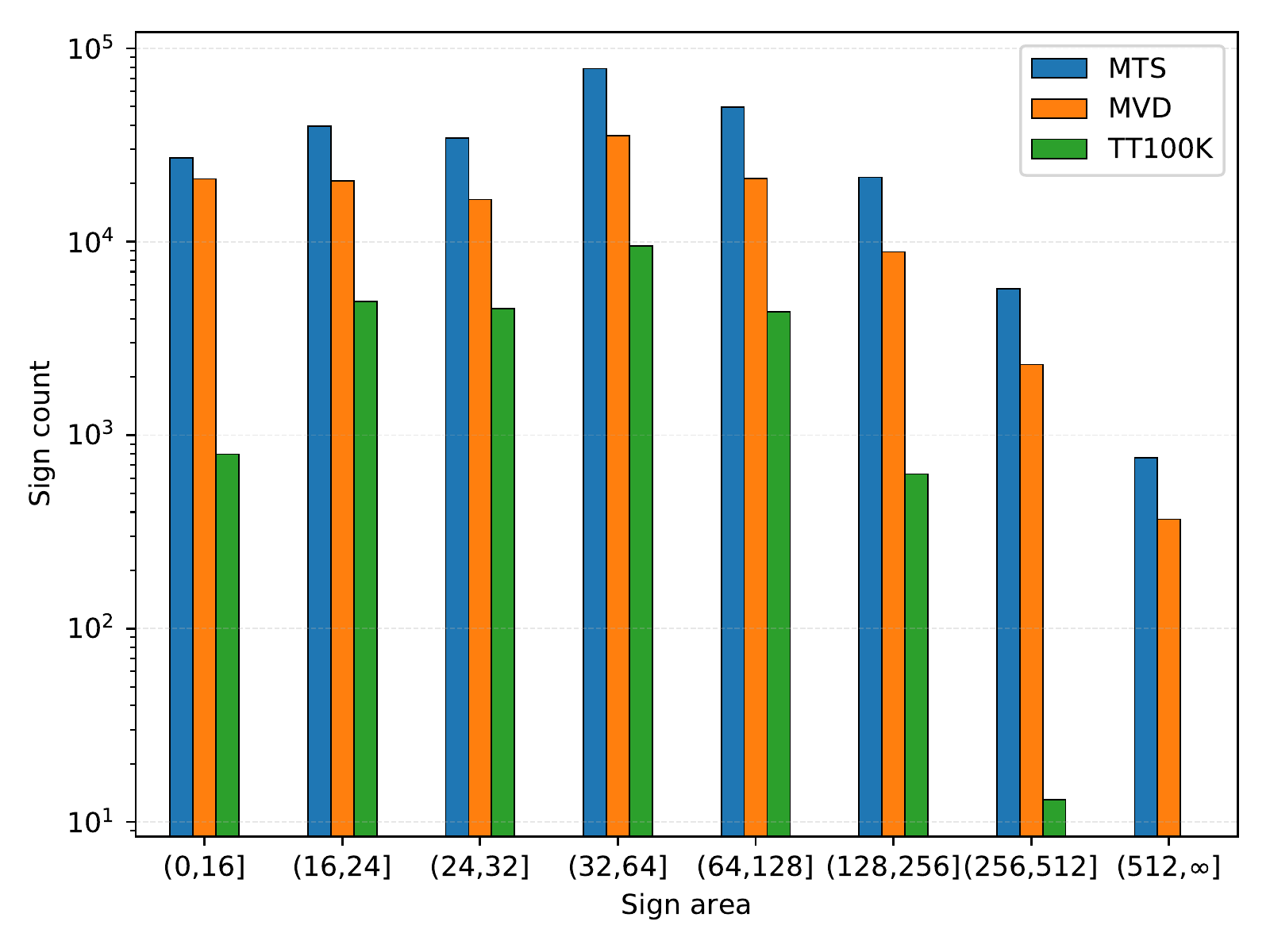}
    \end{subfigure}\hfill%
    \begin{subfigure}[t]{0.33\linewidth}
        \includegraphics[width=\columnwidth]{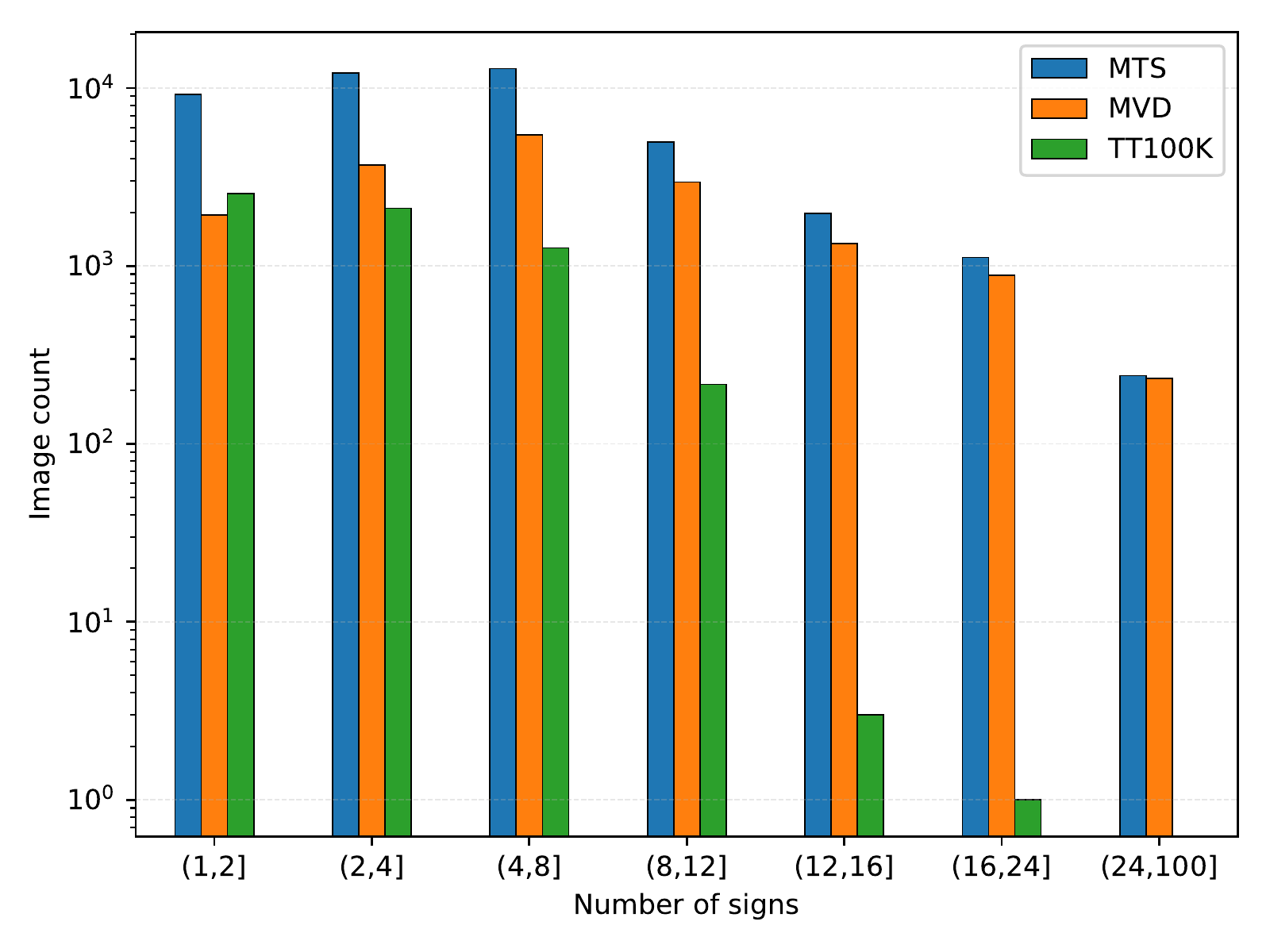}
    \end{subfigure}
    \caption{\emph{Left:} Number of traffic sign classes; \emph{Middle:} Number of signs binned by size; \emph{Right:} Number of images binned by number of traffic sign instances.}  
    \label{fig:sign_distributions}
\end{figure*}

\begin{figure*}
    \centering
        \includegraphics[width=0.9\linewidth]{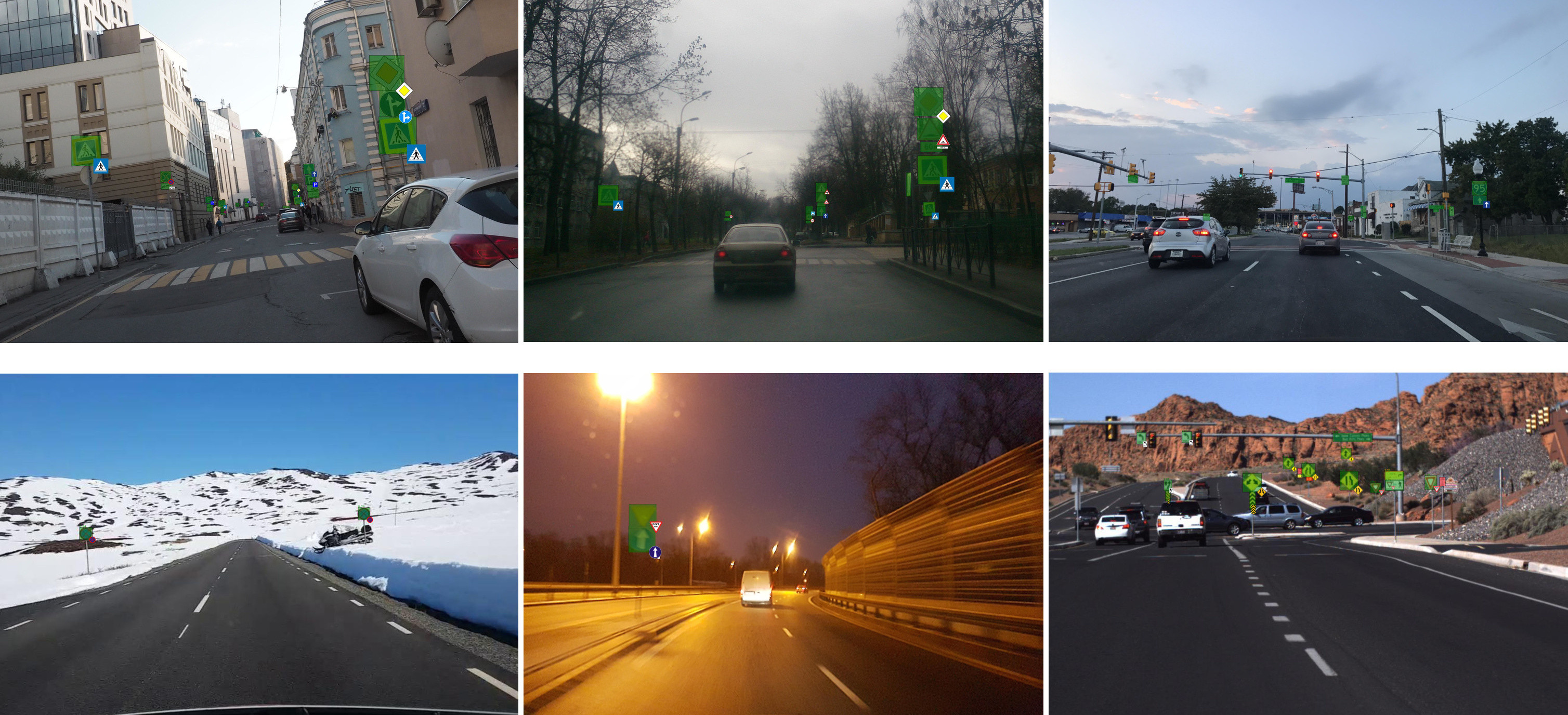}
    \caption{Example images in \mts\ from different geographical locations under varying lighting and weather conditions. Top: Bounding box and class label annotation of traffic signs (green boxes without neighboring template indicts a \emph{other-sign}); Bottom: Results from our detection and classification baseline on the validation set (green colored: true positive, red: missing detections)}
    \label{fig:gt_detections}
\end{figure*}

The fully-annotated set of \mts\ includes a total number of \num{257543} traffic sign bounding boxes out of which \num{82724} have a class label within our traffic sign taxonomy covering \num{313} different traffic sign classes.
The remaining traffic signs sum up as \num{85122} ambiguous signs, \num{23407} directional signs, \num{9141} information signs, \num{3416} highway shields, \num{6451} exterior signs, \num{1917} barrier signs, \num{23468} signs without a selected template, and \num{21897} signs with a template not included by our current taxonomy (but potentially in future releases of the dataset).

The left plot in \cref{fig:sign_distributions} shows a comparison of the traffic sign class distribution between \mts\ and TT100K. Note that MVD is not included here since it does not have labels of traffic sign classes.
\mts\ has approximately twice as many traffic sign classes than TT100K; if we use the definition of a trainable class in~\cite{zhu2016traffic} (which are classes with at least \num{100} traffic sign instances within the dataset) this factor increases to approximately \num{3} between TT100K and \mts.
This difference gets even higher if we consider the instances from the partially annotated set of \mts\ as well.

The plot in the middle of \cref{fig:sign_distributions} compares the areas of signs in terms of pixels in the original resolution of the containing image.
\mts\ covers a broad range of traffic sign sizes with an almost uniform distribution up to \SI[parse-numbers=false]{256^2}{\px}.
MVD has a similar distribution with a lower overall volume.
In comparison to TT100K, \mts\ provides a higher fraction of extreme sizes which poses another challenge for traffic sign detection.

Finally, the plot on the right of \cref{fig:sign_distributions} shows the distribution of images over the number of signs within the image.
Besides the higher volume of images, \mts\ contains a larger fraction of images with a large number of traffic sign instances (\ie $>12$).
One reason for this is that the annotations in \mts\ cover all types of traffic signs, whereas TT100K only contains annotations for very specific types of traffic signs in China. 

\subsection{Annotator Interactions}
\label{sec:human_changes}

\begin{table}[]
\centering
\begin{tabular}{@{}rrlHHHH@{}}
\toprule
\multicolumn{1}{l}{} & \textbf{Count} & \textbf{Mean} & \textbf{Median} & \textbf{25\%} & \textbf{75\%} & \textbf{Max} \\ \midrule
\textit{Images worked on} & \num{52608} & - & - & - & - & - \\
\textit{Signs worked on} & \num{266238} & - & - & - & - & - \\ \midrule
\textit{Originated from detection} & \num{128601} & \num{0.52} & - & - & - & - \\
\textit{IoU with original detection} & - & \num{0.76} & \num{0.84} & \num{0.65} & \num{0.96} & \num{1.0} \\ \midrule
\textit{New signs per image} & - & \num{2.63} & \num{2} & \num{1} & \num{3} & \num{38} \\ \bottomrule
\end{tabular}
\caption{Statistics of manual annotator interactions.}
\label{tab:manual_interaction}
\end{table}

To gather some insights about the work of the annotators, we analyzed their interactions with the bounding boxes fetched from the \emph{Mapillary} API as described in \cref{sec:annot_process} and present the results in \cref{tab:manual_interaction}.
We found that \SI{52}{\percent} of the bounding boxes annotated in \mts\ originated from an automatic detection already present.
However, when comparing the final boxes within \mts\ to the original ones, we find an overlap of only \SI{76}{\percent} in terms of IoU, proving the improvement of detection accuracy.
Additionally, we found that the annotators on average added approximately \num{3} completely new bounding boxes that were missing before in each image.


\section{Traffic Sign Detection}
\label{sec:detection}

The first task defined on \mts\ is detecting traffic signs as bounding boxes without inferring the specific class labels. The goal is to predict a set of bounding boxes with corresponding confidence scores as traffic signs. 

\begin{description}[style=unboxed,leftmargin=0cm,listparindent=\parindent]

    \item[Metrics.]
    Given a set of detections with estimated scores for each image, we first compute the matching between the detections and annotated ground truth within each image separately.
    A detection can be successfully matched to a ground truth if their Jaccard overlap ($\mathrm{IoU}$)~\cite{everingham2015pascal} is $>0.5$; if multiple detections match the same ground truth, only the detection with the highest score is a match while the rest is not (\emph{double detections}); each detection will only be matched to one ground truth bounding box with the highest overlap.
    
    Having this matching indicator (TP vs.\ FP) for every detection, we define define average precision~(\AP) similar to COCO~\cite{lin2014microsoft} (\ie $\AP^{\mathrm{IoU=}0.5}$ which resembles \AP\ definition of PASCAL~VOC~\cite{everingham2015pascal}) and compute precision as a function of recall by sorting the matching indicators by their corresponding detection confidence scores in descending order and accumulate the number of TPs and FPs.
    \AP\ is then defined as the area under the curve of this step function.
    Additionally, we follow~\cite{lin2014microsoft} and compute \AP\ for traffic signs of different scales: \APs, \APm, and \APl\ refer to \AP\ computed for boxes with area $a<32^2$, $32^2<a<96^2$, and $a > 96^2$, respectively.
    
    \item[Baseline and Results.]
    In \cref{tab:localization_results}, we show experimental results using a Faster R-CNN based detector~\cite{ren2015faster} with FPN~\cite{lin2017feature} and residual networks~\cite{he2016deep} as the backbone.
    
    During training we randomly sample crops of size $\num{1000}\times\num{1000}$ at full resolution instead of down-scaling the image to avoid vanishing of small traffic signs, as traffic signs can be very small in terms of pixels and \mts\ covers traffic signs from a broad range of scales in different image resolutions.
    We use a batch size of 16 distributed over 4 GPUs during training for the ResNet50 models; for the ResNet101 version, we use batches of size 8. 
    Unless stated otherwise, we train using stochastic gradient descent~(SGD) with an initial learning rate of \num{1e-2} and lower the learning rate when the validation error plateaus.
    For inference, we down-scale the input images such that their larger side does not exceed a certain number of pixels (either \SI{2048}{\px} or \SI{4000}{\px}) or operate on full resolution if the original image is smaller. 
    
    Besides training on \mts, we conduct transfer-learning experiments on TT100K and MVD\footnote{We convert the segmentation of \emph{traffic-sign--front} instances to bounding boxes by taking the minimum and maximum in the x, y axes. Note that this conversion can be inaccurate if signs are occluded by other objects.} to test the generalization properties of the proposed dataset.
    We use the same baseline as for the \mts\ experiments and train it on both datasets, once with ImageNet initialization and once with \mts\ initialization.
    The models trained with ImageNet initialization are trained to convergence. To ensure a fair comparison, we fine-tune only for half the number of epochs when initializing with \mts\ weights.
    The results in \cref{tab:localization_results} show that \mts\ pre-training boosts detection performance by a large margin on both datasets regardless of the input resolution.
    This is a clear indication for the generalization qualities of \mts.
    
\end{description}

\begin{table*}[]
\renewcommand*{\thefootnote}{\fnsymbol{footnote}}
\small
\centering
\begin{tabular}{@{}lllll|llll@{}}
\toprule
& \multicolumn{4}{c}{Max 4000px} & \multicolumn{4}{c}{Max 2048px}   \\ \cmidrule(l){2-9} 
 & \AP & \APs & \APm & \APl &  \AP & \APs & \APm & \APl  \\ \midrule
\multicolumn{9}{c}{\textbf{MTSD}} \\ \midrule
ResNet50 FPN & \num{87.3} & \num{73.03} & \num{91.91} & \num{93.56} & \num{80.22} & \num{52.31} & \num{88.87} & \num{94.73} \\
ResNet101 FPN & \num{88.44} & \num{74.00} & \num{92.14} & \num{93.70} & \num{81.80} & \num{56.55} & \num{89.22} & \num{94.82} \\ \midrule
\multicolumn{9}{c}{\textbf{TT100K}} \\ \midrule
\cite{zhu2016traffic}~multi-scale\footnotemark[1] & \num{91.79} & \num{84.56} & \num{96.40} & \num{92.60}  & - & - & - & -\\ \midrule
ResNet50 FPN & - & - & - & - & \num{91.27} & \num{84.01} & \num{95.87} & \num{90.13} \\
\ + pre-trained on \mts\ & \- &  - & - & -  & \textbf{\num{97.60}} {\smaller (+\num{6.33})}  & \num{93.13} & \num{99.03} & \num{98.44} \\ \midrule
\multicolumn{9}{c}{\textbf{MVD (traffic signs)}} \\ \midrule
ResNet50 FPN & \num{72.90} & \num{46.60} & \num{79.93} & \num{85.42} & \num{64.00} & \num{30.70} & \num{75.28} & \num{86.50} \\
\ + pre-trained on \mts\ & \textbf{\num{76.31}} {\smaller (+\num{3.41})} & \num{51.00} & \num{83.49} & \num{88.33} & \textbf{\num{68.29}} {\smaller (+\num{4.29})}  & \num{33.60} & \num{79.45} & \num{89.53} \\ \bottomrule
\end{tabular}
\caption{Detection results on \mts, TT100K and MVD. Numbers in brackets refer to absolute improvements when pre-training on \mts\ in comparison to ImageNet. \protect\footnotemark[1] They evaluate using multi-scale inference with scales $0.5$, $1$, $2$, and $4$.}
\label{tab:localization_results}
\end{table*}

\begin{table}[]
\small
\centering
\begin{tabular}{@{}lllll@{}}
\toprule
 & \mAP & \mAPs & \mAPm & \mAPl \\ \midrule
\multicolumn{5}{c}{\textbf{MTSD}} \\ \midrule
FPN50 + classifier & \num{81.1} & \num{69.4} & \num{85.0} & \num{87.2} \\
FPN101 + classifier & \num{83.4} & \num{76.4} & \num{85.8} & \num{87.3} \\ \midrule
\multicolumn{5}{c}{\textbf{TT100K}} \\ \midrule
\cite{zhu2016traffic}~multi-scale & \num{81.6} & \num{68.3} & \num{86.5} & \num{85.7} \\ \midrule
FPN50 + classifier & \textbf{\num{89.9}} {\smaller(+\num{8.3})} & \num{83.9} & \num{93.0} & \num{84.3} \\
+ \emph{det  pre-trained} & \textbf{\num{93.4}} {\smaller(+\num{11.8})} & \num{88.2} & \num{94.8} & \num{93.6} \\
+ \emph{cls pre-trained} & \textbf{\num{95.7}} {\smaller(+\num{14.1})} & \num{91.3} & \num{96.9} & \num{96.7} \\ \bottomrule
\end{tabular}
\caption{Simultaneous detection and classification results. The numbers in brackets are absolute improvements over~\cite{zhu2016traffic}. \emph{det~pre-trained} and \emph{cls~ pre-trained} refer to experiments with additionally \mts\ pre-trained detector and classifier, respectively.}
\label{tab:classification_results}
\end{table}

\section{Simultaneous Detection and Classification}

\begin{figure}
    \centering
    \includegraphics[width=\linewidth]{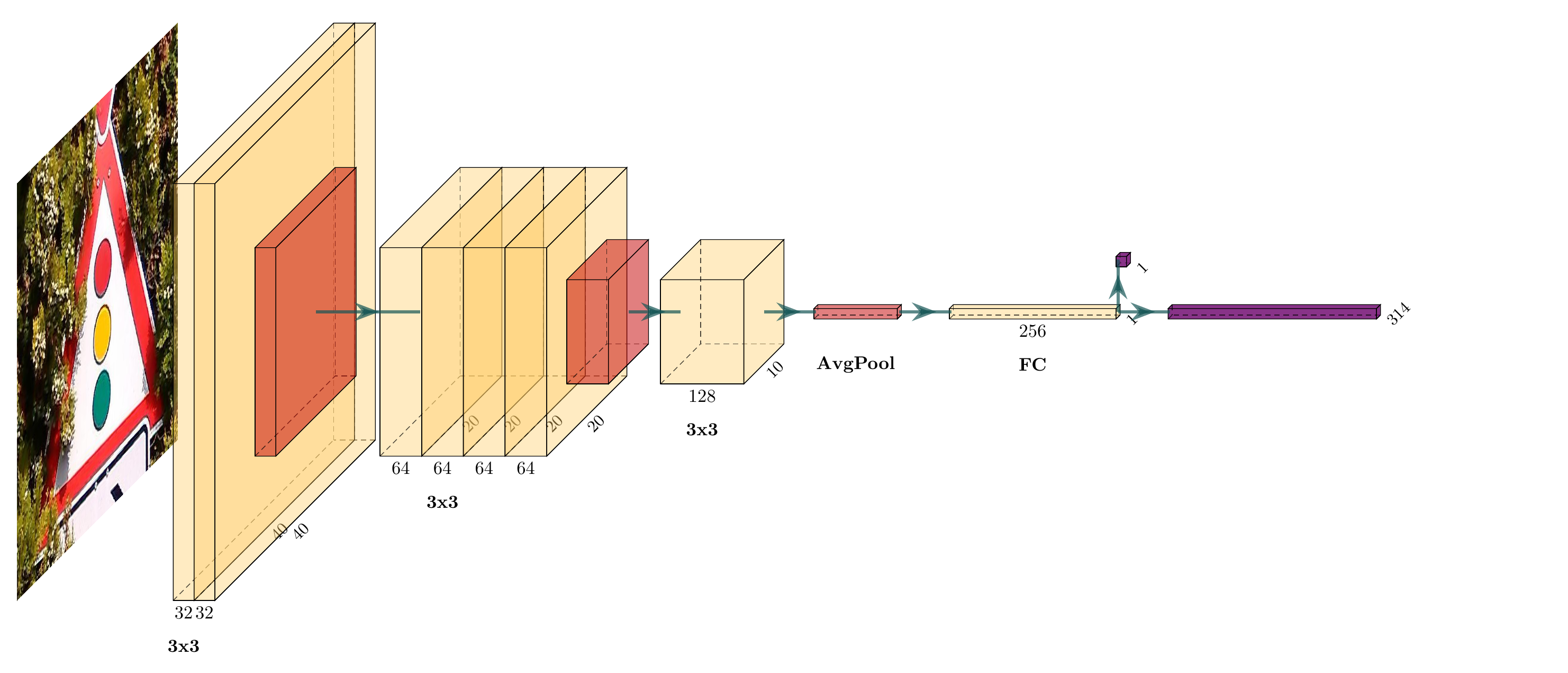}
    \caption{Network architecture of the baseline classifier.}
    \label{fig:classifier_architecture}
\end{figure}

\begin{figure}
    \centering
    \includegraphics[width=\linewidth]{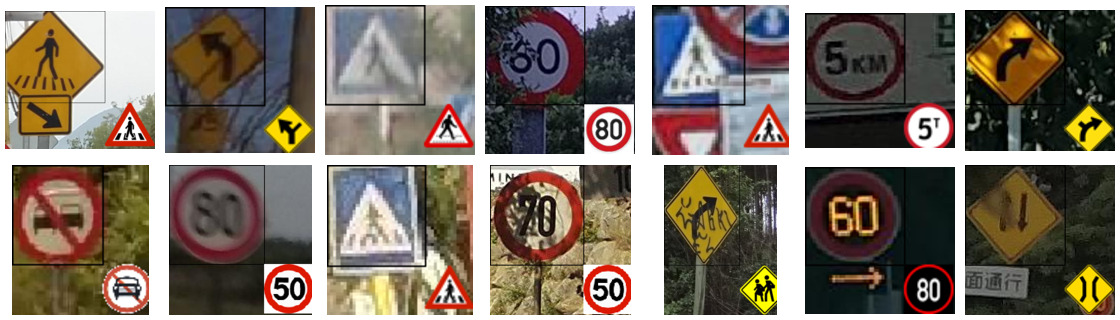}
    \caption{Failure cases of the baseline classification network on \mts.}
    \label{fig:failure_cases}
\end{figure}

The second task on \mts\ is simultaneous detection and classification of traffic signs (\ie multi-class detection).
It extends the detection task to additionally predict a class label for each traffic sign instance that is in our taxonomy.
For the traffic signs that do not have a label within our taxonomy we introduce a general class \emph{other-sign}.

\begin{description}[style=unboxed,leftmargin=0cm,listparindent=\parindent]

    \item[Metric.]
    The main metric for this task is mean average precision~(\mAP) over all \num{313} classes; per-class \AP\ is calculated as described in \cref{sec:detection}.
    The matching between predicted and ground truth boxes is done in a binary way by ignoring the class label.
    After that, we filter out all \emph{other-sign} ground-truth instances and detections since we do not want to evaluate on this general class.
    
    \item[Baseline.]
    A trivial baseline for this task would be to extend the binary detection baseline from \cref{sec:detection} to the multi-class setting by adding a \num{314}-way classification head.
    However, preliminary experiments showed that a straight-forward training of such a network does not yield acceptable performance.
    We hypothesize that this is due to \begin{enumerate*}[label=(\arabic*)]
    \item scale issues for small signs before RoI pooling and, 
    \item under-represented class variation within the training batches given that the majority of traffic sign instances are \emph{other-sign}.
    \end{enumerate*}
    
    To overcome the scale issue and to have better control over batch statistics during training, we opted for a two-stage architecture with using our binary detector in the first stage and a decoupled shallow classification network in the second stage. This form of decoupling has been shown to improve the detection and recognition accuracy \cite{cheng2018revisiting}. 
    The classification network consists of seven $3 \times 3$ convolutions (each followed by batch normalization) with 2 max-pooling layers after the $2^{nd}$ and $6^{th}$ convolution layer.
    The last convolution is followed by spatial average pooling and a fully-connected stage resulting in a \num{314}-way classification head with softmax activation ($313$ and \emph{other-sign}) and a single sigmoid activation for foreground/background classification. The network architecture is depicted in \cref{fig:classifier_architecture}.
    
    We use image crops predicted by the detector (both foreground and background) together with crops from the ground-truth as input during training and optimize the network using cross-entropy loss.
    To balance the distribution of traffic sign classes in a batch, we uniformly sample \num{128} different classes with 3 samples each class and add another \num{128} background crops per batch.
    We train the network with SGD for \num{30} epochs starting with a learning rate of \num{1e-2}, lowered by a factor of \num{0.1} after \num{10} and \num{20} epochs.
    
    \item[Results.]
    We show the results of our baseline in \cref{tab:classification_results}.
    Our classifier in combination with ResNet101 binary detector reaches \num{83.4} \mAP\ over all \num{313} classes; the ResNet50 variant is only about \num{2} points lower.
    \cref{fig:gt_detections} shows visual examples of our baseline's predictions and \cref{fig:failure_cases} shows typical failure cases of the classification network.
    
    To verify our baseline, we train with the same setup on TT100K and compare the results with the baseline in~\cite{zhu2016traffic}\footnote{We convert their results to the format used by \mts\ and evaluate using our metrics.}.
    Our two-stage approach outperforms their baseline by \num{8.3} points, even though the performances of the binary detectors are similar (see \cref{tab:localization_results}).
    This validates that the decoupled classifier (even with a very shallow network) is able to yield good results.
    Moreover, the accuracy is improved further after pre-training the classifier (and the detector) on \mts\ before fine-tuning it on TT100K, which further validate the generalization effectiveness of the \mts.  
    
\end{description}

\section{Conclusion}

In this work, we have introduced \mts, a large-scale traffic sign benchmark dataset that includes \num{100}$K$ images with full and partial bounding-box annotations, covering \num{313} traffic sign classes from all over the world. \mts\ is the most diverse traffic sign benchmark dataset in terms of geographical locations, scene characteristics, and traffic sign classes.  We have shown in baseline experiments that decoupling detection and fine-grained classification yields superior results on previous traffic sign datasets. Additionally, in transfer-learning experiments, we show that \mts\ facilitates fine-tuning and improves accuracy substantially for traffic sign datasets in a narrow domain.  

We see \mts\ as the first step to drive the research efforts towards solving fine-grained traffic sign detection and classification at a global scale. With the partial annotated dataset, we would also like to pave the way for further research in semi-supervised learning. In the future, we would like to extend the dataset towards a complete traffic sign taxonomy globally. To achieve this, we see the potential of applying zero-shot learning to efficiently model the semantic and appearance attributes of traffic sign classes. With the global taxonomy built, we can optimize the performance further with hierarchical classification~\cite{redmon2017yolo9000, yan2015hd}.

{\small
\bibliographystyle{ieee}
\bibliography{mly}
}

\end{document}


\title{The Mapillary Traffic Sign Dataset for Detection and Classification on a Global Scale\\Supplementary Material}

\author{Christian Ertler \and Jerneja Mislej \and Tobias Ollmann \and Lorenzo Porzi \and Gerhard Neuhold \and Yubin Kuang\\
{\tt\small \emph{\{christian, jerneja, tobias, lorenzo, gerhard, yubin\}}@mapillary.com}
}

\maketitle


\section{Scene Classification for Image Selection}
An important requirement during image selection for \mts\ was to ensure high diversity of images with different image properties. 
Since the frequency of occurrence of certain traffic sign classes can be very different depending on the scene, we trained a neural network to predict the scene classes of the images and used the predicted labels to guide the image selection in order to diversify the scene classes in the final dataset. To train the scene classification network, we have used a subset of the scene classes of the BDD100K dataset~\cite{yu2018bdd100k}.
After filtering BDD100K for images that have either the \emph{residential}, \emph{highway}, or \emph{city street} class label, we trained a ResNet50~\cite{ren2015faster} that was pre-trained on ImageNet with a cross-entropy loss using stochastic gradient descent~(SGD).
The network was trained to convergence with an initial learning rate of \num{1e-2} which was reduced by a factor of \num{0.1} until validation accuracy plateaued. 

\begin{figure}
    \centering
    \includegraphics[width=\linewidth]{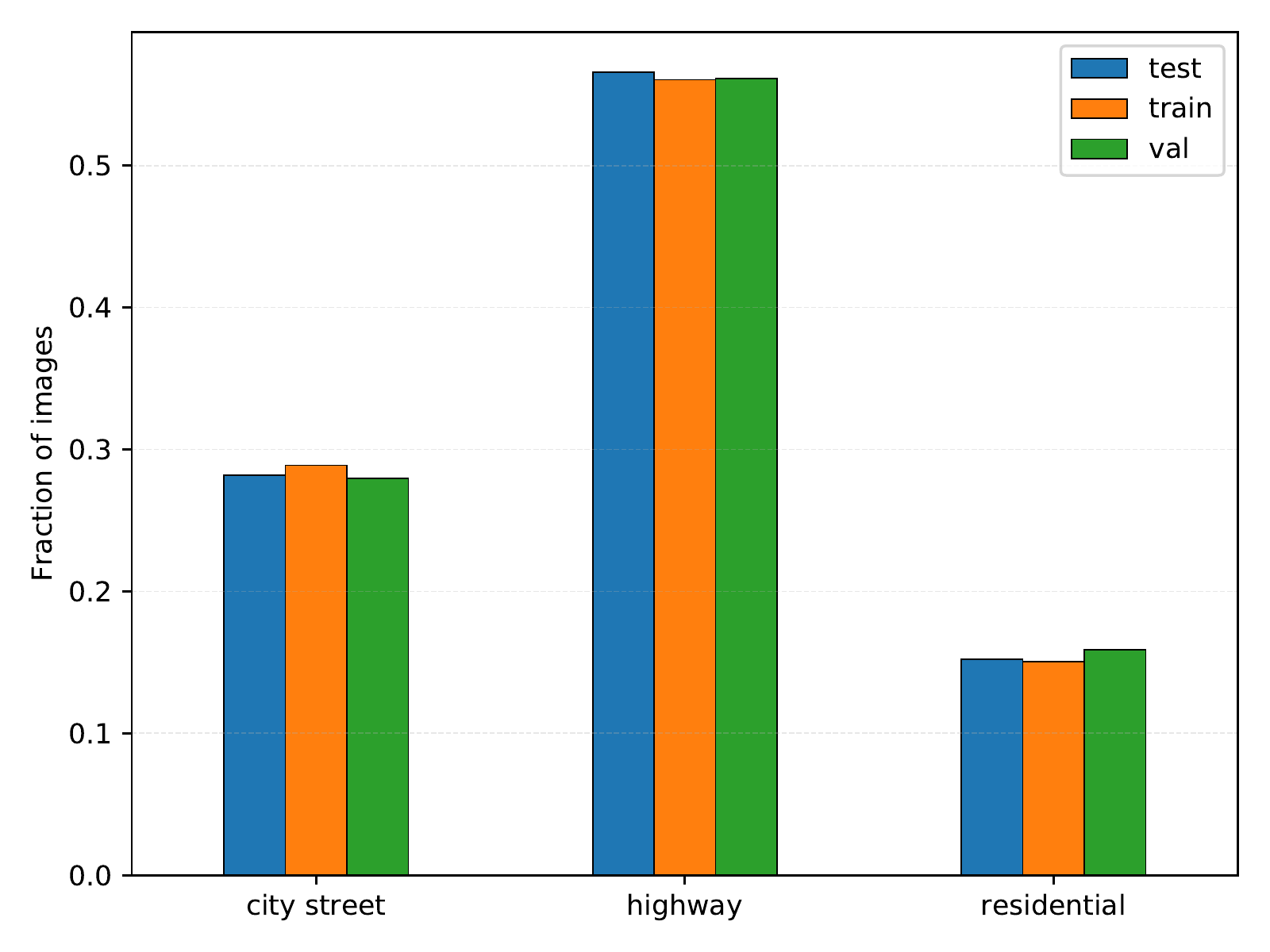}
    \caption{Distribution of scene categories within \mts\ as predicted by our scene prediction network.}
    \label{fig:scene_distribution}
\end{figure}

\cref{fig:scene_distribution} shows the distribution of scene classes within the supervised set of \mts\ according to predictions of this model as targeted during our greedy image selection scheme.
We opted for a uniform distribution after treating \emph{city street} and \emph{residential} as a single class, since we found that these two classes (as annotated in BDD100K) are not always clearly distinguishable even for human. Given the large number of candidate images, this weakly-supervised image selection scheme facilitated increasing the diversity in scene classes.

\section{Template Proposal Network}

\begin{figure*}
    \centering
    \includegraphics[width=\linewidth]{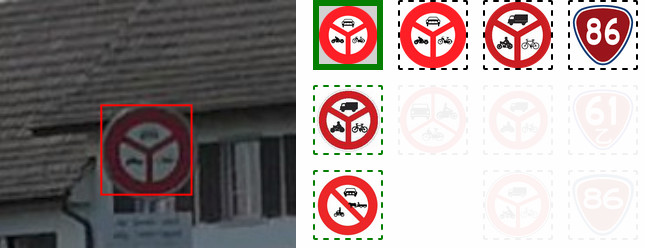}
    \caption{The classification UI used by the annotators. The traffic sign to be annotated is shown with its bounding box on the left. On the right, one can see the current selection (green bounding box in the \nth{1} column) as well as the proposed templates.
    Each column starting with the second one shows a proposed template group based on the similarity of the real image crop and the templates as predicted by our proposal network; if the similarity is below a certain threshold, the templates are grayed out (starting from \nth{5} column in this example).
    The other templates in the \nth{1} column show proposals based on the currently selected template; note how this enables the annotator to find even more similar templates that are not proposed based on the image crop.}
    \label{fig:classification_ui}
\end{figure*}

As mentioned in Section~2.2 of the main paper, we used a neural network to predict similarities between sampled crops and grouped template images in order to assist the annotators in choosing a valid template in the annotation process.
The predicted similarities were used to propose template images for each sign in a similarity-ordered way.
Without such a mechanism, it would be extremely time-consuming for the annotators to handle the large set of different template images that are available.

We use a metric learning approach~\cite{bellet2013survey} to train a 3-layer network (similar to but shallower than the baseline classification network in the main paper) to learn a function $f(x): \mathbb{R}^d \to \mathbb{R}^k$ that maps a $d$-dimensional input vector to a $k$-dimensional embedding space.
In our case, $x$ are input images encoded as vectors of size $d = 40 \times 40 \times 3$ and $k = 128$.
We train the network with a contrastive loss~\cite{hadsell2006dimensionality} such that the cosine similarity
\begin{equation}
    \Sim(x_1, x_2) = \frac{x_1^{{\mathsmaller T}} x_2}{\left\Vert x_1 \right\Vert_{2} \left\Vert x_2 \right\Vert_{2}}
\end{equation}
between two embedding vectors $x_1$ and $x_2$ with corresponding group labels $\hat{y}_1$, $\hat{y}_2$ should be high if the samples are within the same template group, whereas the similarity should be lower than a margin $m$ if the samples are from different groups:
\begin{equation}
    \mathcal{L} =
    \begin{cases}
        1 - \Sim(x_1, x_2), &\quad\text{if } \hat{y}_1 = \hat{y}_2\\
        \max\left[0, \Sim(x_1, x_2) - m\right] &\quad\text{else}
    \end{cases}.
\end{equation}
We choose $m = 0.2$ and train the network using a generated training set by blending our traffic sign templates to random background images after scaling, rotating and sheering it by a reasonable amount.

Note that the goal of the model is not necessarily to predict the correct class in terms of the most similar template but to have the matching template together with similar ones at least within the top-k predictions.
In this way, the annotator can browse the template groups either ordered by similarity to the traffic sign crop under question, or ordered by the similarity between a selected template image and other templates.
The latter ordering allows to browse through the template images in a semantically meaningful way if a matching template is not proposed in the first place.
Further, we want to point out that this approach allowed us to add new missing templates to the UI on demand without the need of training data or re-training of the proposer network.
\cref{fig:classification_ui} shows a screenshot of the user interface using the described network to propose templates.

Besides the proposer based navigation, we additionally provided a text-based template search. This was necessary for cases where the proposer failed to provide good templates.

\section{Partial Annotation}
In this section, we elaborate on how we automatically generated the partially annotated images using a structure from motion pipeline. For each fully-annotated image within the training set of \mts, we query for a set of neighboring images from Mapillary that locate within a pre-defined distance to form a image cluster. Then, we recovered the relative camera poses between images in the cluster using a pipeline based on OpenSfM~\cite{opensfm}. To create tentative correspondences between annotated signs and automatic detections (by Mapillary) in the neighboring images, we rely on the class labels \ie a pair of signs with the same labels form a tentative correspondence. With such tentative correspondences, we further triangulate the 3D positions of the signs~\cite{hartley2003multiple} and vote for the most geometrically feasible correspondences based on the estimated relative camera poses. Here, we triangulate the traffic signs as 3D points with the centers of corresponding 2D bounding boxes.

To this end, we have established geometrically and semantically consistent correspondences between the annotated signs and automatic detections. The correspondences are then utilized to generate the partial annotated dataset as described in main paper by propagating the human verified class labels of the corresponding traffic sign instances in the fully annotated training set to the automatically generated ones.

\section{Qualitative Examples}

In the following we show additional examples of annotated \mts\ images in \cref{sec:mts_annots}.
Further, we show results of our transfer learning experiments on TT100K~\cite{zhu2016traffic} and MVD~\cite{neuhold2017mapillary} in \cref{sec:transfer_results}.
For qualitative comparisons of detections, we make sure that we choose score thresholds so that either recall or precision are comparable.

\subsection{Examples in \mts}
\label{sec:mts_annots}

\begin{figure*}
    \centering
    \begin{subfigure}[t]{0.495\linewidth}
        \includegraphics[width=\linewidth]{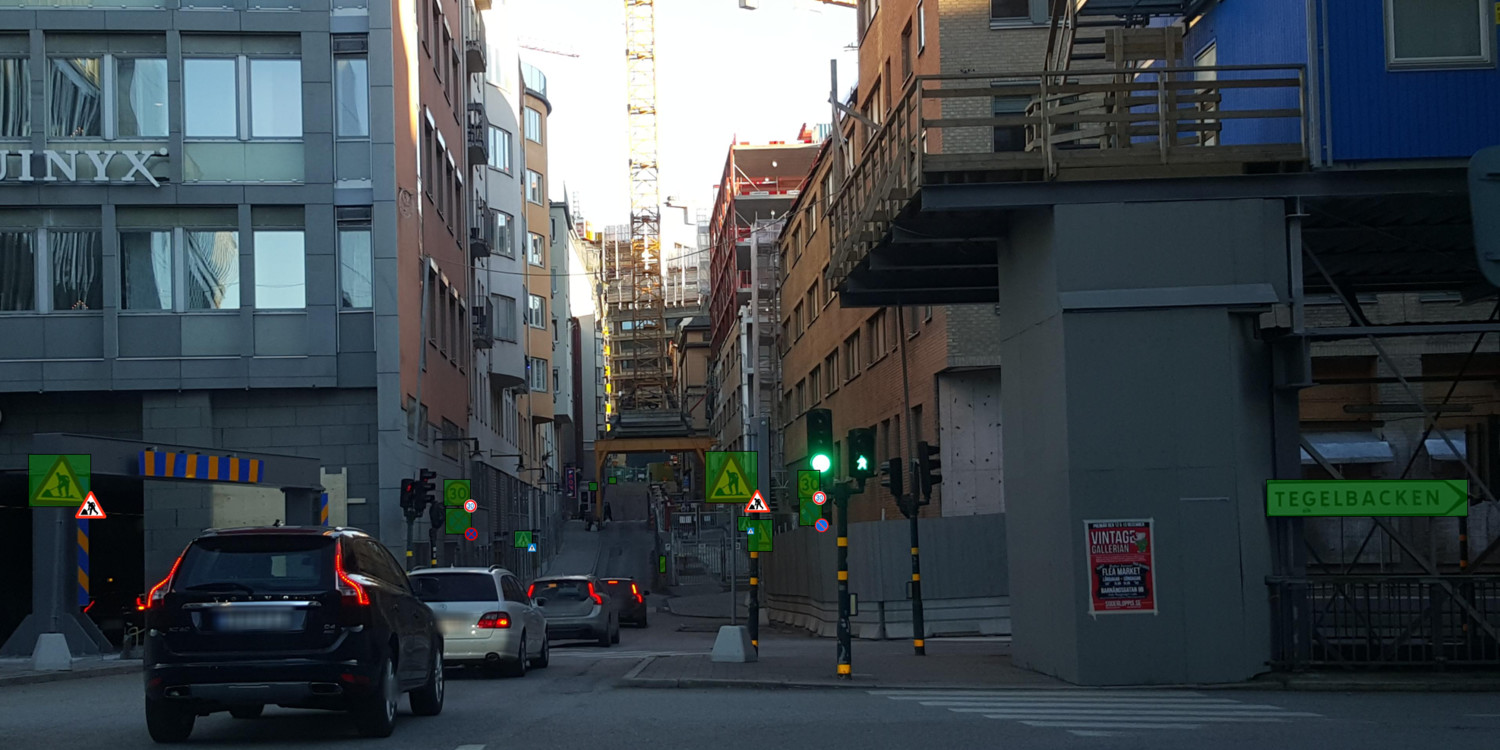}
    \end{subfigure}\hfill%
    \begin{subfigure}[t]{0.495\linewidth}
        \includegraphics[width=\linewidth]{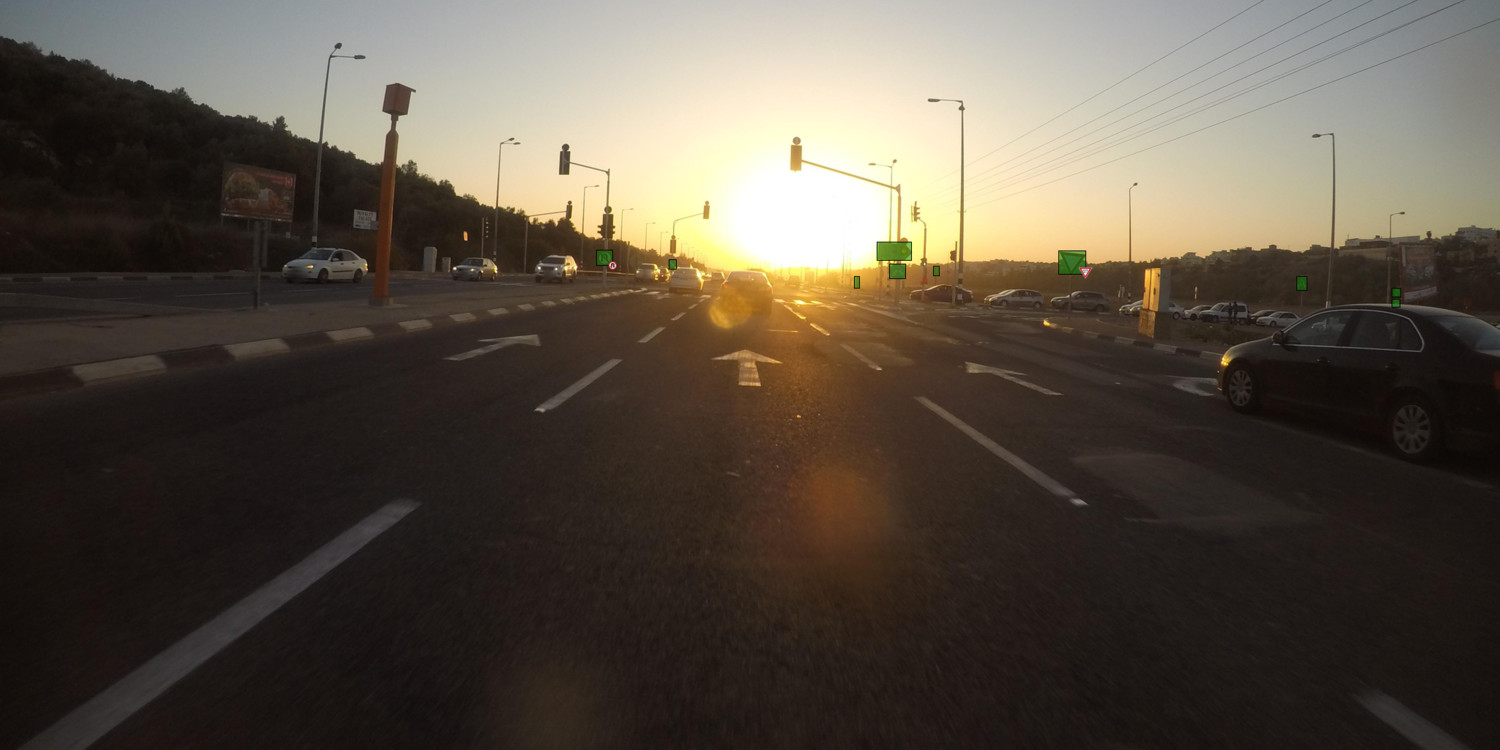}
    \end{subfigure}\\[0.01\linewidth]%
    \begin{subfigure}[t]{0.495\linewidth}
        \includegraphics[width=\linewidth]{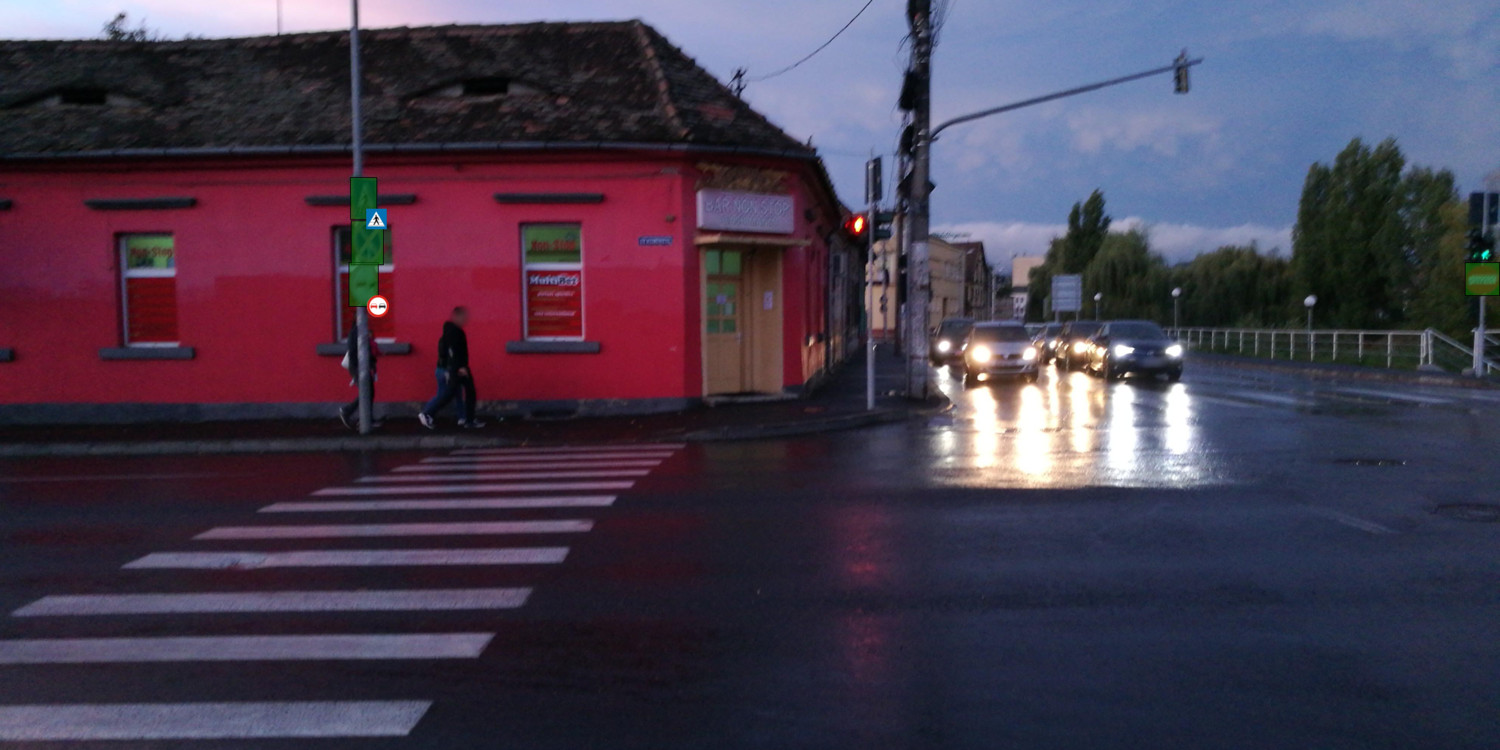}
    \end{subfigure}\hfill%
    \begin{subfigure}[t]{0.495\linewidth}
        \includegraphics[width=\linewidth]{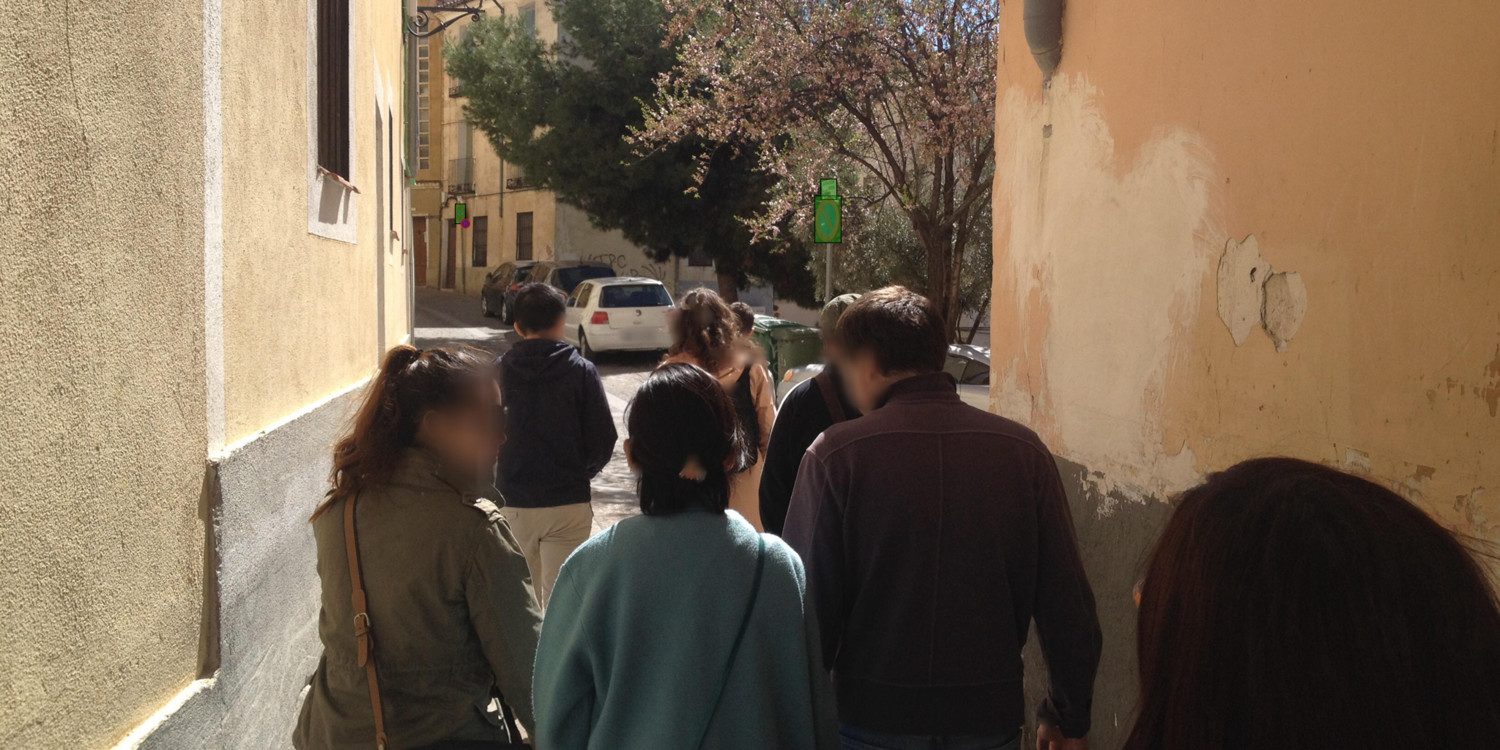}
    \end{subfigure}\\[0.01\linewidth]%
    \begin{subfigure}[t]{0.495\linewidth}
        \includegraphics[width=\linewidth]{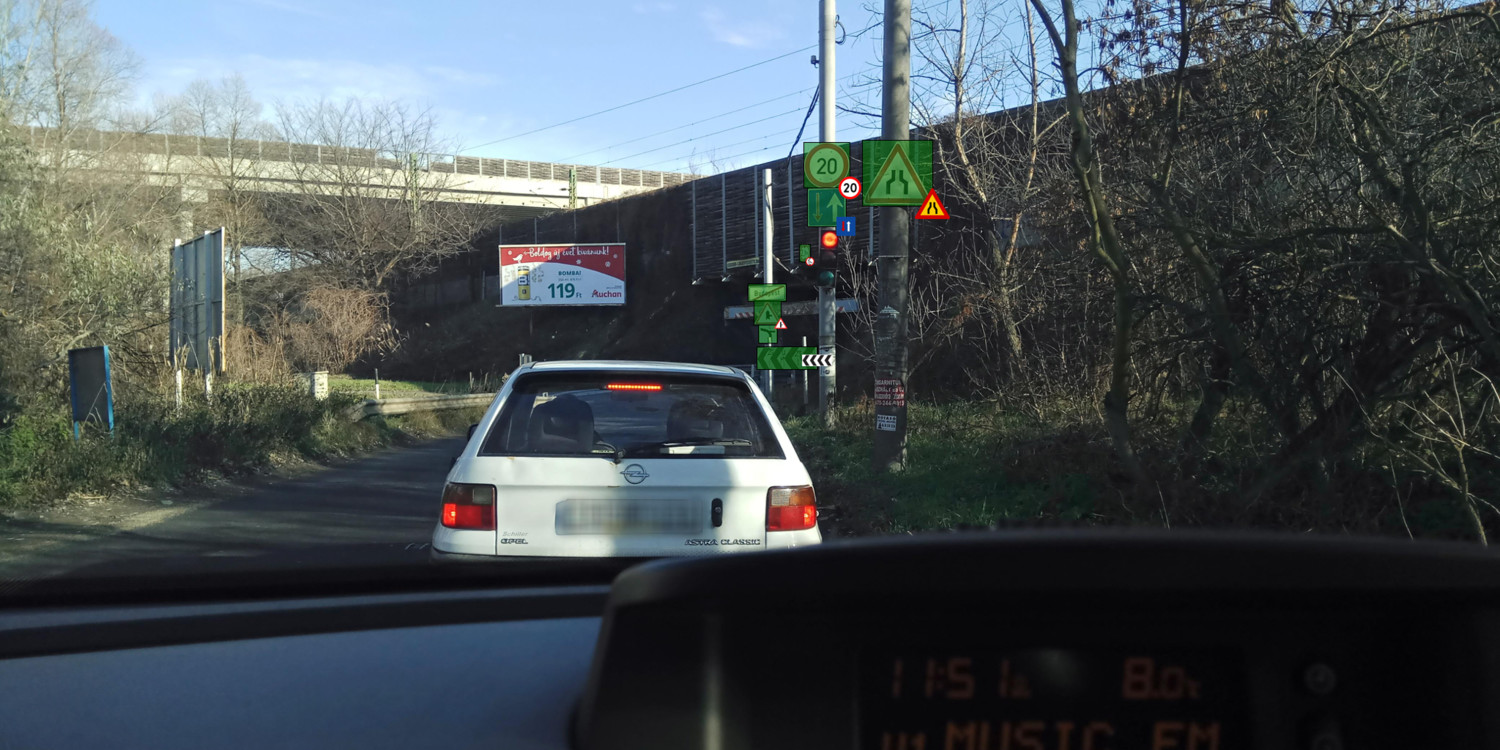}
    \end{subfigure}\hfill%
    \begin{subfigure}[t]{0.495\linewidth}
        \includegraphics[width=\linewidth]{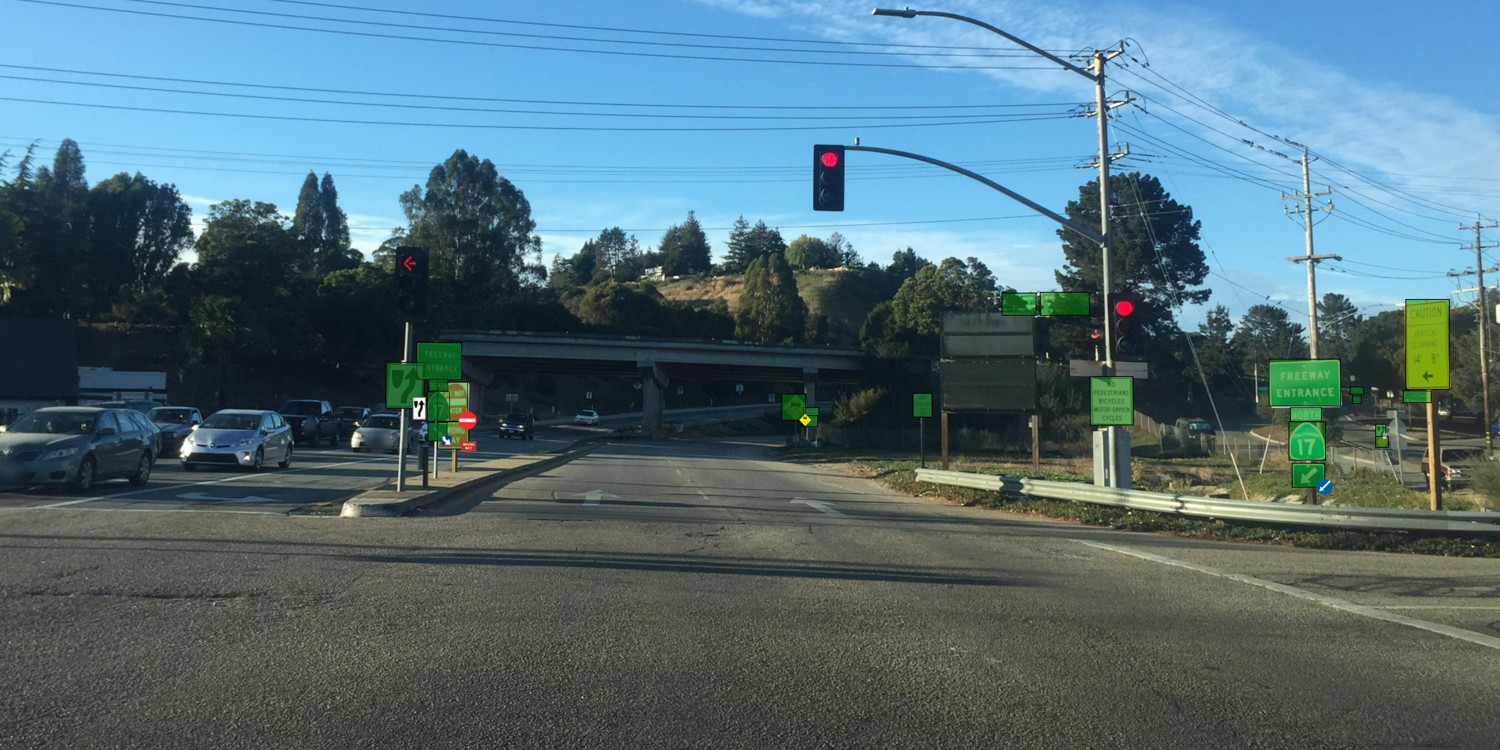}
    \end{subfigure}\\[0.01\linewidth]%
    \begin{subfigure}[t]{0.495\linewidth}
        \includegraphics[width=\linewidth]{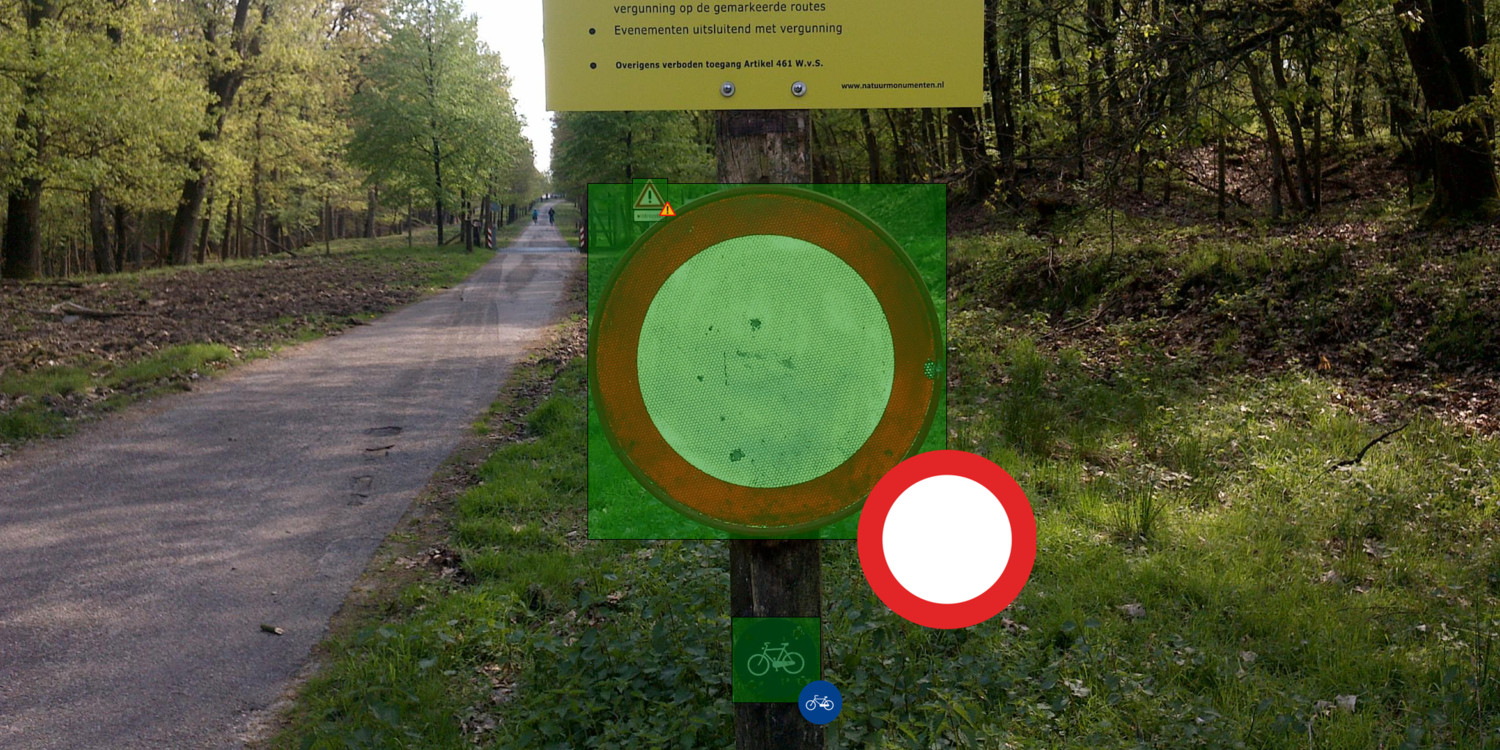}
    \end{subfigure}\hfill%
    \begin{subfigure}[t]{0.495\linewidth}
        \includegraphics[width=\linewidth]{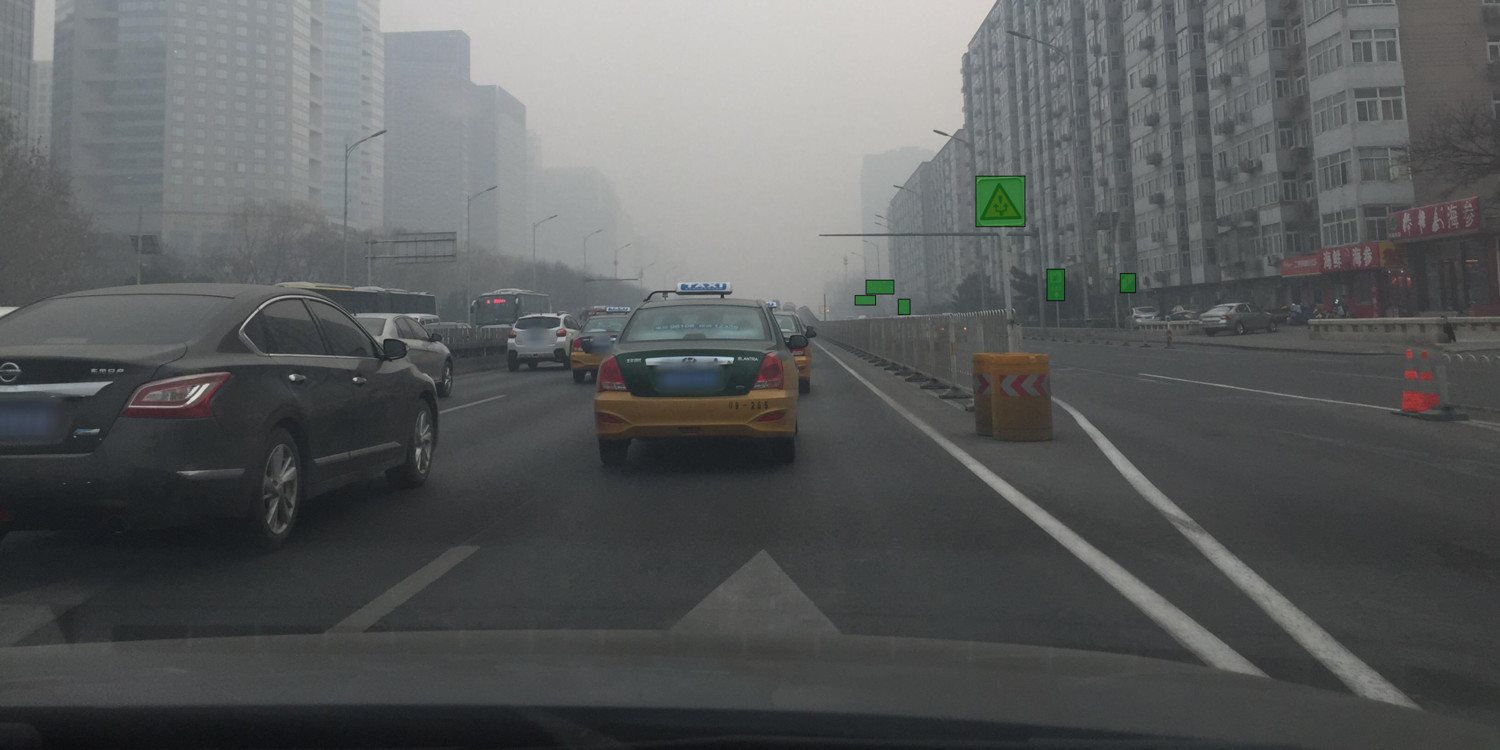}
    \end{subfigure}
    \caption{Examples of annotated images from the \mts\ training set, covering diverse lighting and weather conditions}
    \label{fig:training_samples}
\end{figure*}

We show some examples of annotated images from the \mts\ training set in \cref{fig:training_samples}. \mts\ covers a broad range of capture settings including cities, highways, residential areas, and rural areas with different lighting and weather conditions from varying view points.
This variety makes \mts\ the most diverse traffic sign dataset available.

\subsection{Impact of Transfer Learning}
\label{sec:transfer_results}

\begin{figure*}
    \centering
    \begin{subfigure}[t]{0.495\linewidth}
        \includegraphics[width=\linewidth]{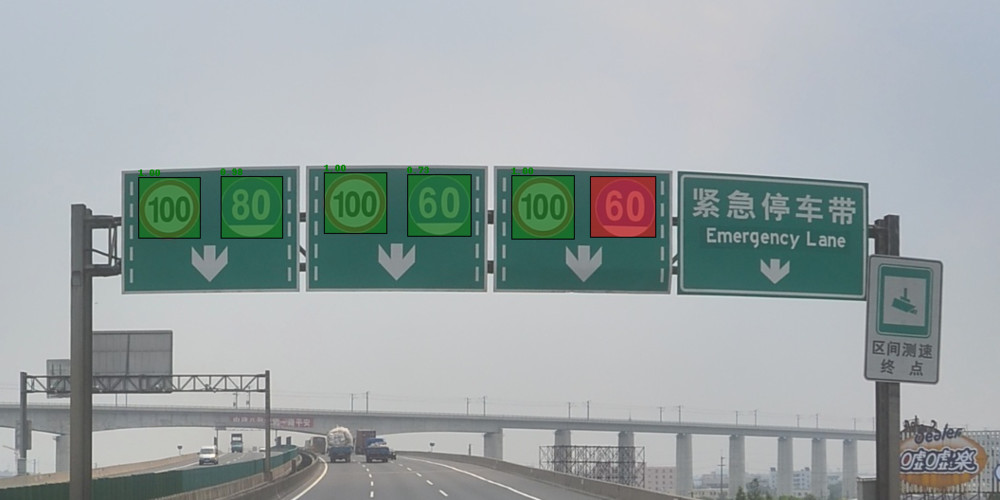}
    \end{subfigure}\hfill%
    \begin{subfigure}[t]{0.495\linewidth}
        \includegraphics[width=\linewidth]{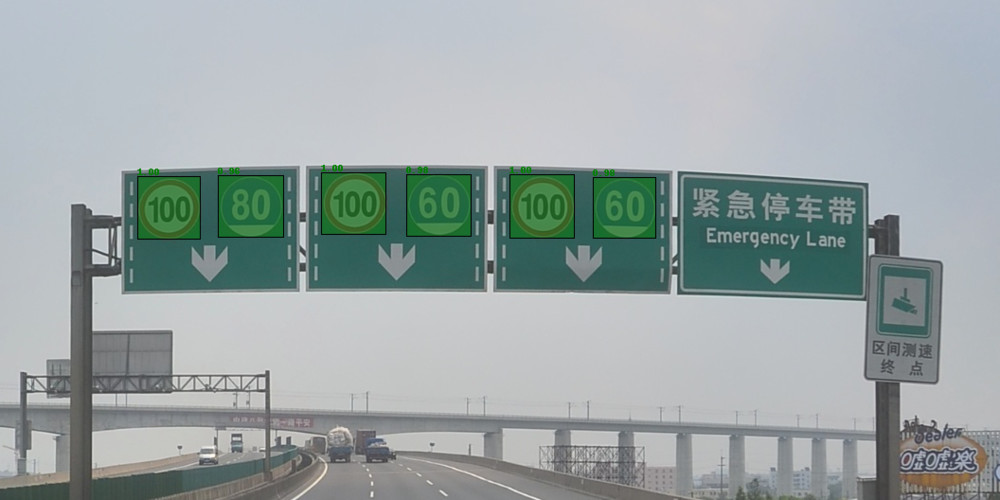}
    \end{subfigure}\\[0.01\linewidth]%
    \begin{subfigure}[t]{0.495\linewidth}
        \includegraphics[width=\linewidth]{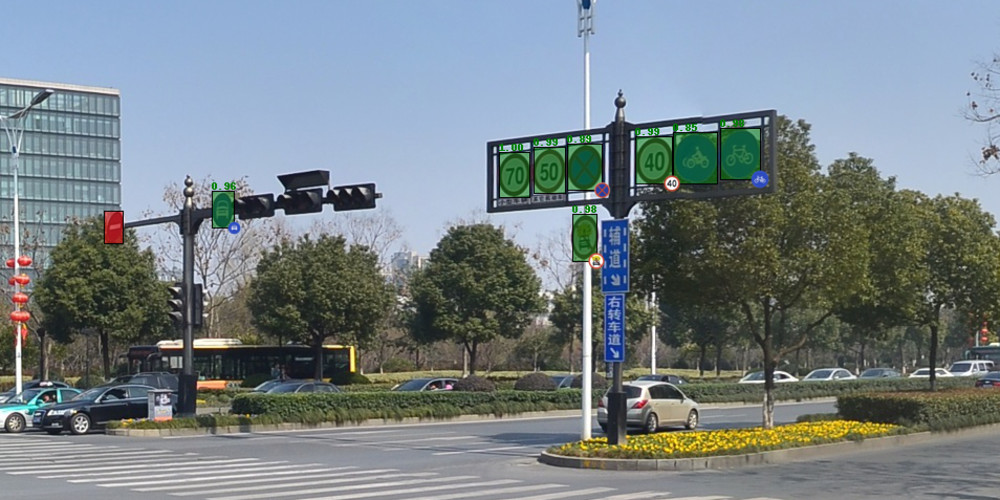}
    \end{subfigure}\hfill%
    \begin{subfigure}[t]{0.495\linewidth}
        \includegraphics[width=\linewidth]{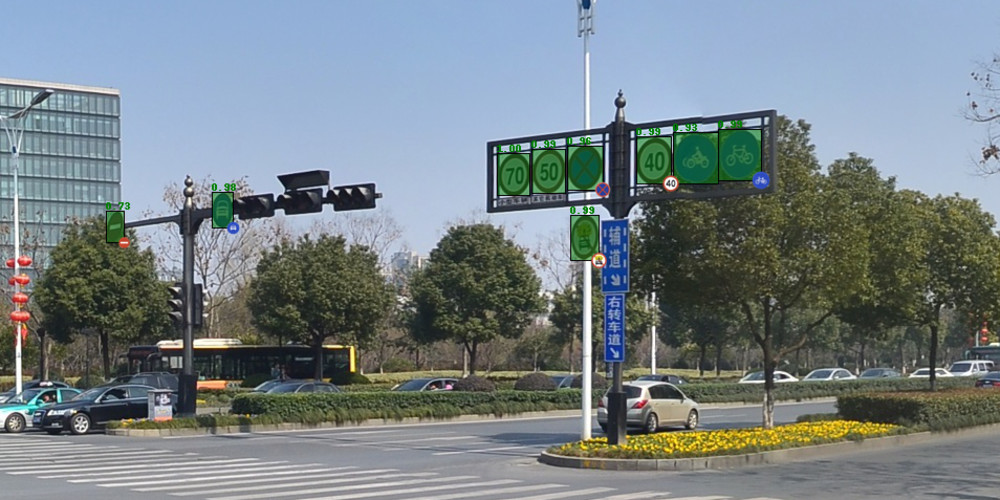}
    \end{subfigure}\\[0.01\linewidth]%
    \begin{subfigure}[t]{0.495\linewidth}
        \includegraphics[width=\linewidth]{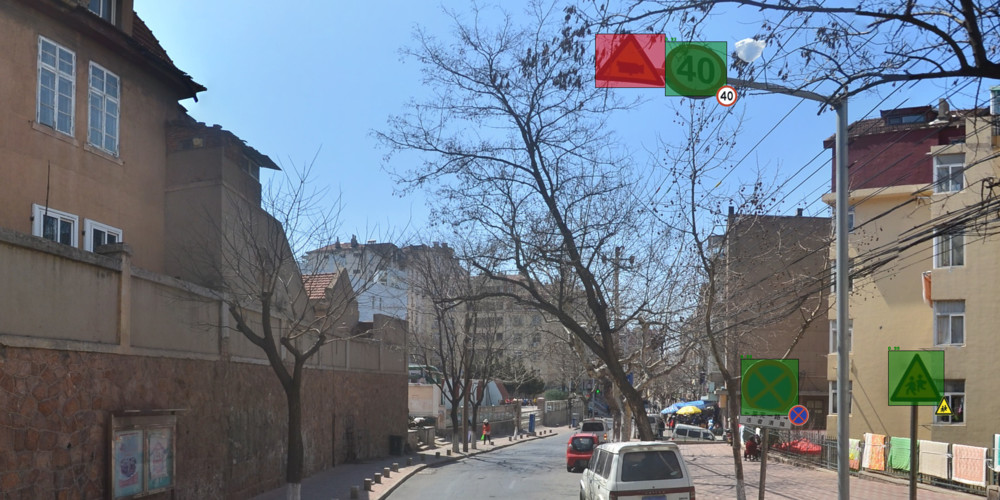}
    \end{subfigure}\hfill%
    \begin{subfigure}[t]{0.495\linewidth}
        \includegraphics[width=\linewidth]{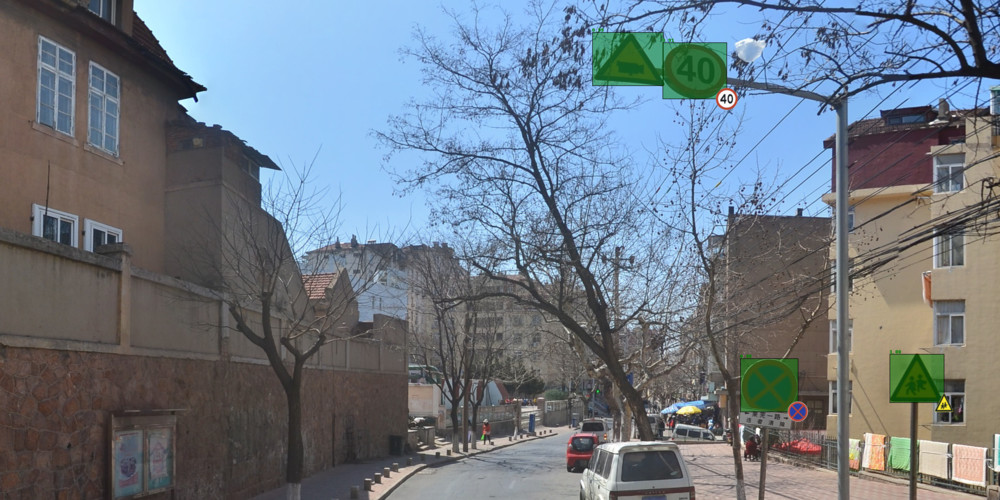}
    \end{subfigure}\\[0.01\linewidth]%
    \begin{subfigure}[t]{0.495\linewidth}
        \includegraphics[width=\linewidth]{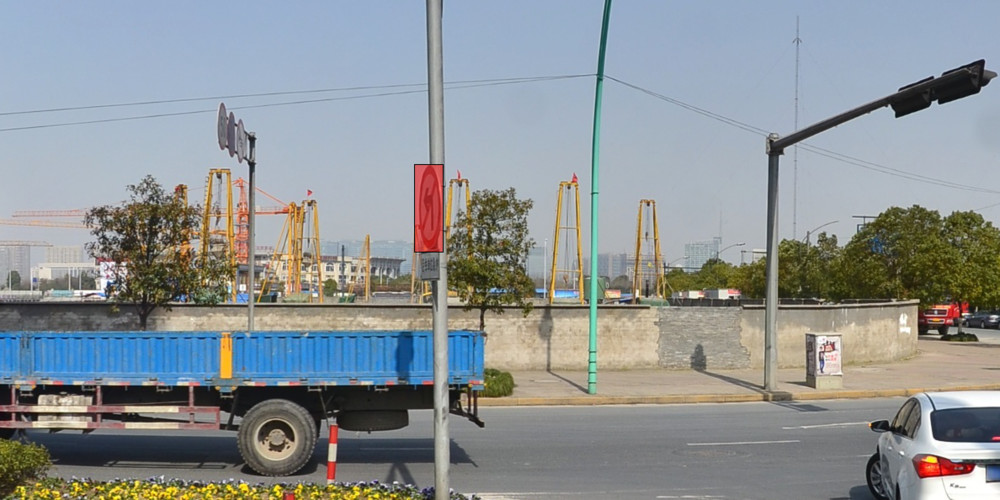}
    \end{subfigure}\hfill%
    \begin{subfigure}[t]{0.495\linewidth}
        \includegraphics[width=\linewidth]{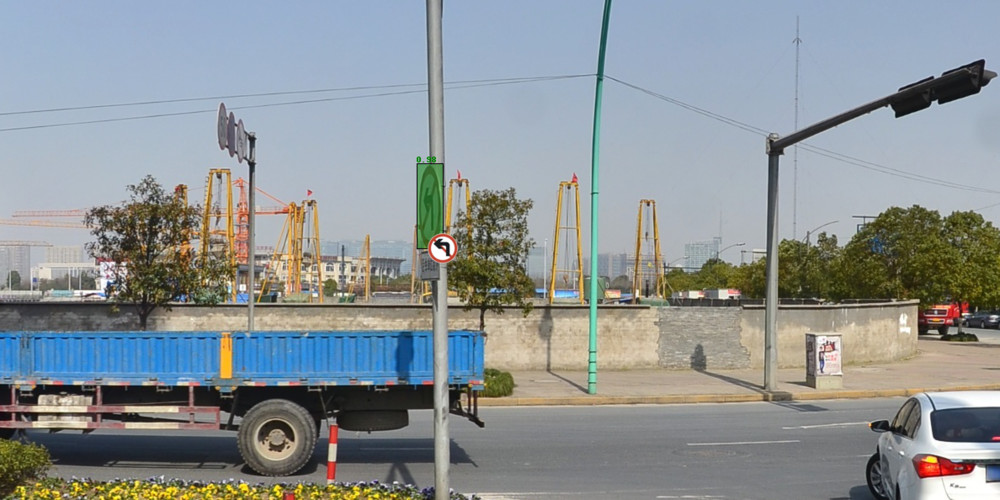}
    \end{subfigure}
    \caption{Qualitative comparisons between our baseline trained on TT100K only~(left), and our baseline pre-trained on \mts\ and fine-tuned on TT100K~(right). The score thresholds are chosen such that both models operate on the same level of precision.}
    \label{fig:tt100k_comp}
\end{figure*}

To illustrate the gains of our baseline on TT100K by pre-training the model on \mts, we show qualitative comparisons of detections in \cref{fig:tt100k_comp}.
The model pre-trained on \mts\ is able to detect more traffic signs in many cases while preserving a high precision.
For fair qualitative comparison, both models operate on the same level of precision (\num{0.95}), however, the model pre-trained on \mts\ achieves a higher recall (\num{0.91} \emph{vs.}\ \num{0.81}).

\begin{figure*}
    \centering
    \begin{subfigure}[t]{0.495\linewidth}
        \includegraphics[width=\linewidth]{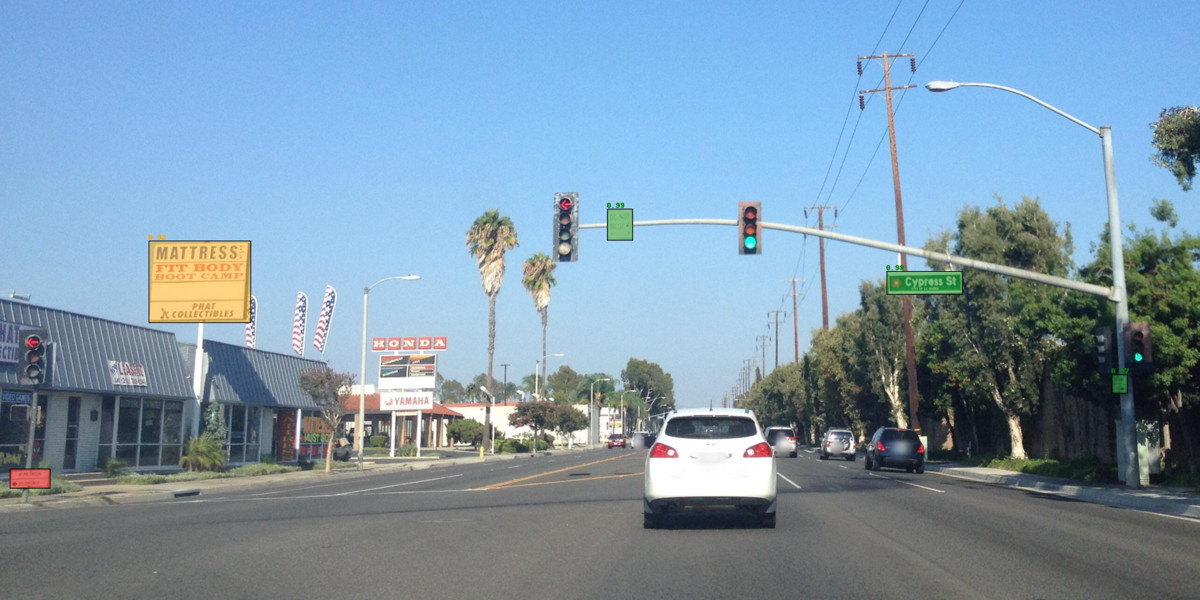}
    \end{subfigure}\hfill%
    \begin{subfigure}[t]{0.495\linewidth}
        \includegraphics[width=\linewidth]{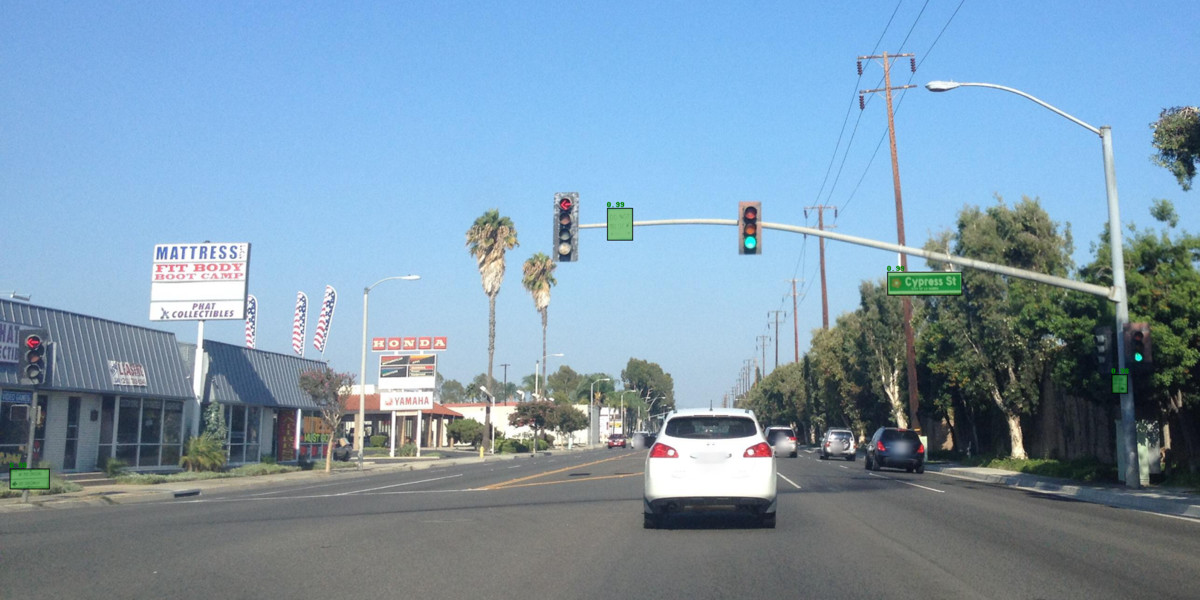}
    \end{subfigure}\\[0.01\linewidth]%
    \begin{subfigure}[t]{0.495\linewidth}
        \includegraphics[width=\linewidth]{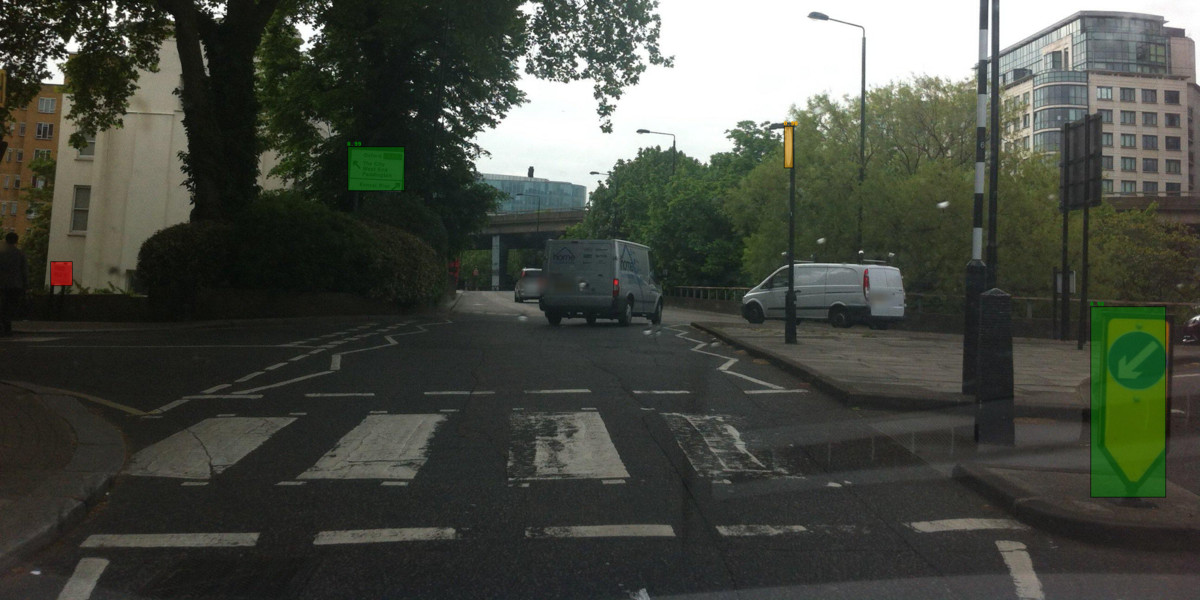}
    \end{subfigure}\hfill%
    \begin{subfigure}[t]{0.495\linewidth}
        \includegraphics[width=\linewidth]{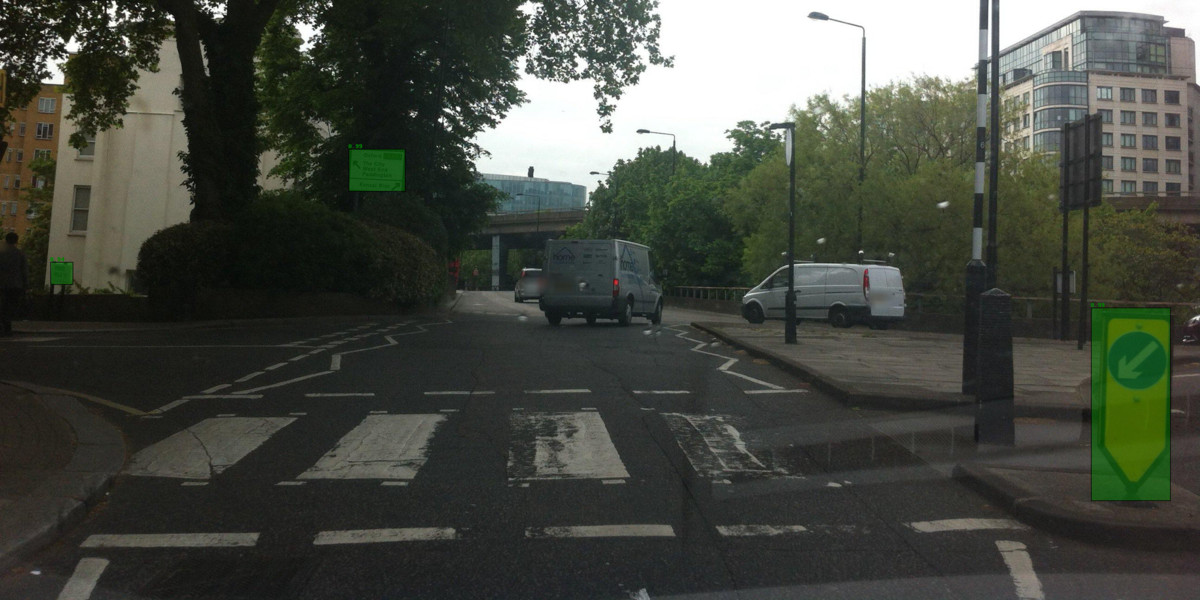}
    \end{subfigure}\\[0.01\linewidth]%
    \begin{subfigure}[t]{0.495\linewidth}
        \includegraphics[width=\linewidth]{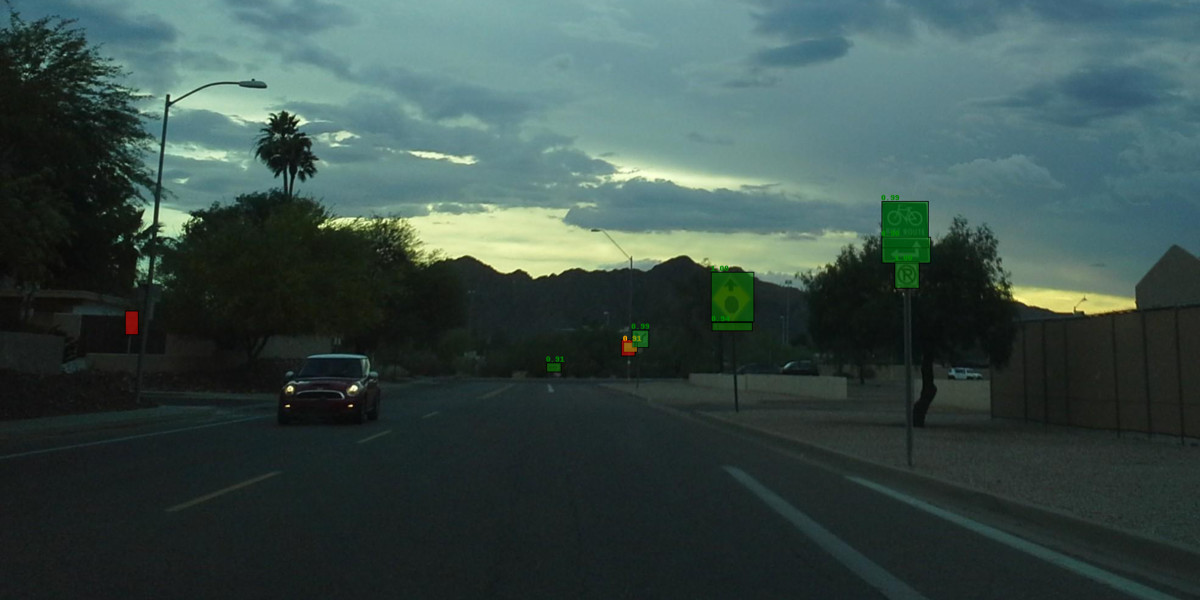}
    \end{subfigure}\hfill%
    \begin{subfigure}[t]{0.495\linewidth}
        \includegraphics[width=\linewidth]{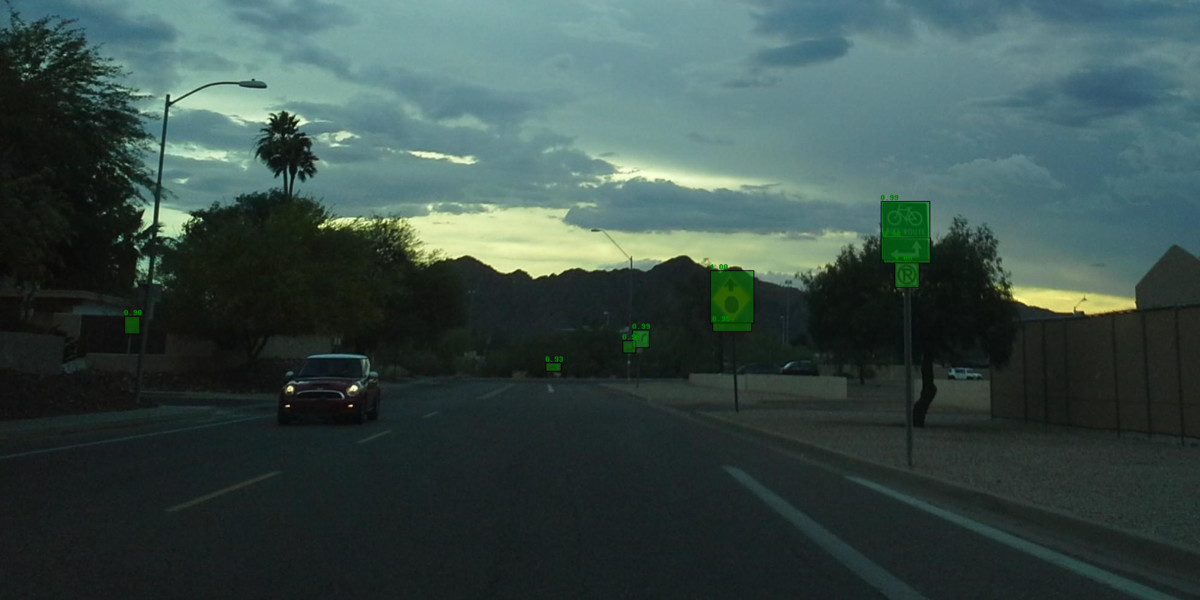}
    \end{subfigure}\\[0.01\linewidth]%
    \begin{subfigure}[t]{0.495\linewidth}
        \includegraphics[width=\linewidth]{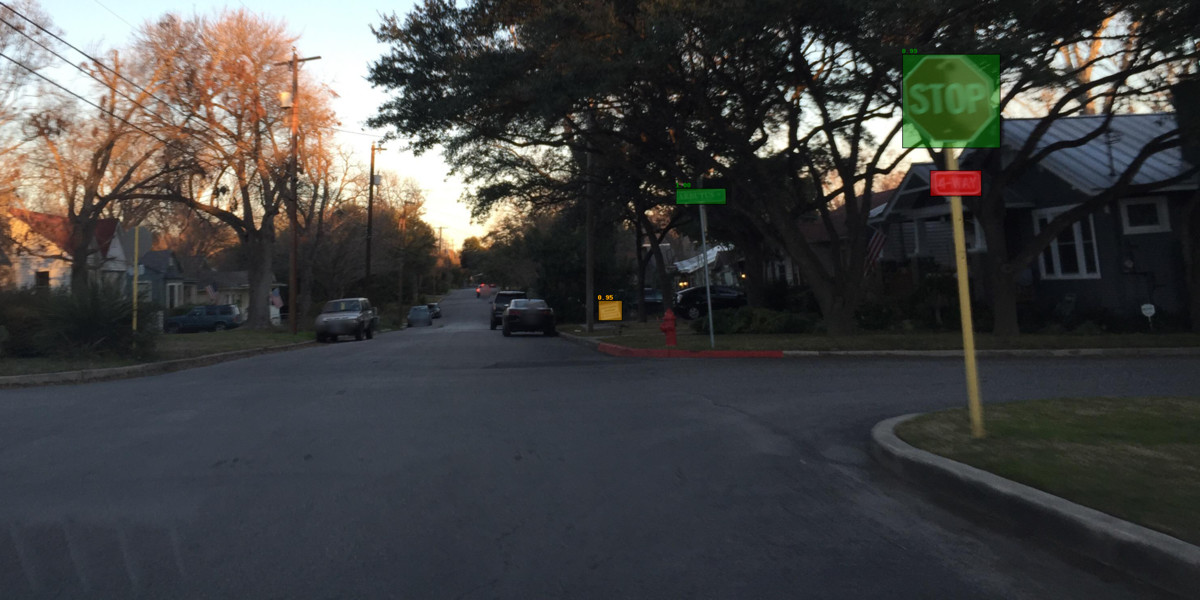}
    \end{subfigure}\hfill%
    \begin{subfigure}[t]{0.495\linewidth}
        \includegraphics[width=\linewidth]{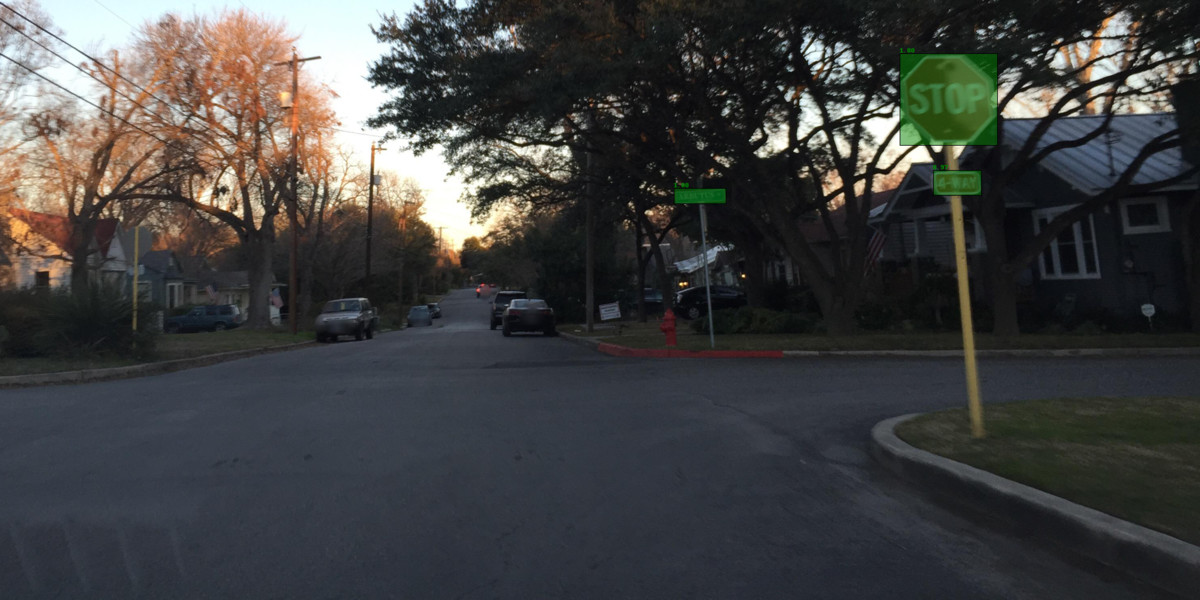}
    \end{subfigure}\\[0.01\linewidth]%
    \begin{subfigure}[t]{0.495\linewidth}
        \includegraphics[width=\linewidth]{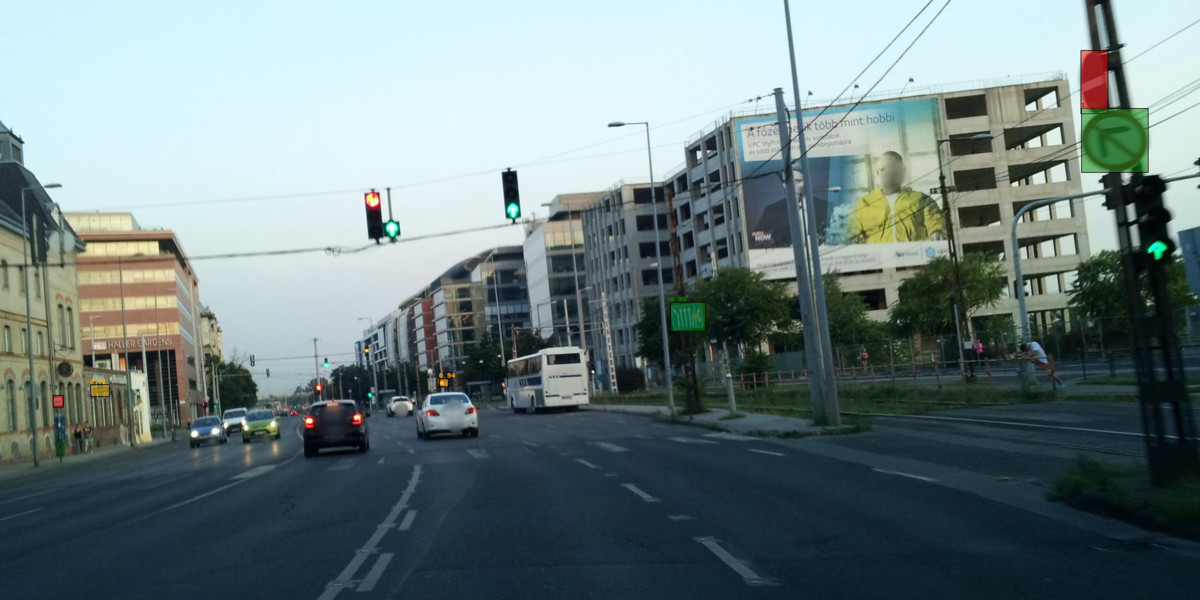}
    \end{subfigure}\hfill%
    \begin{subfigure}[t]{0.495\linewidth}
        \includegraphics[width=\linewidth]{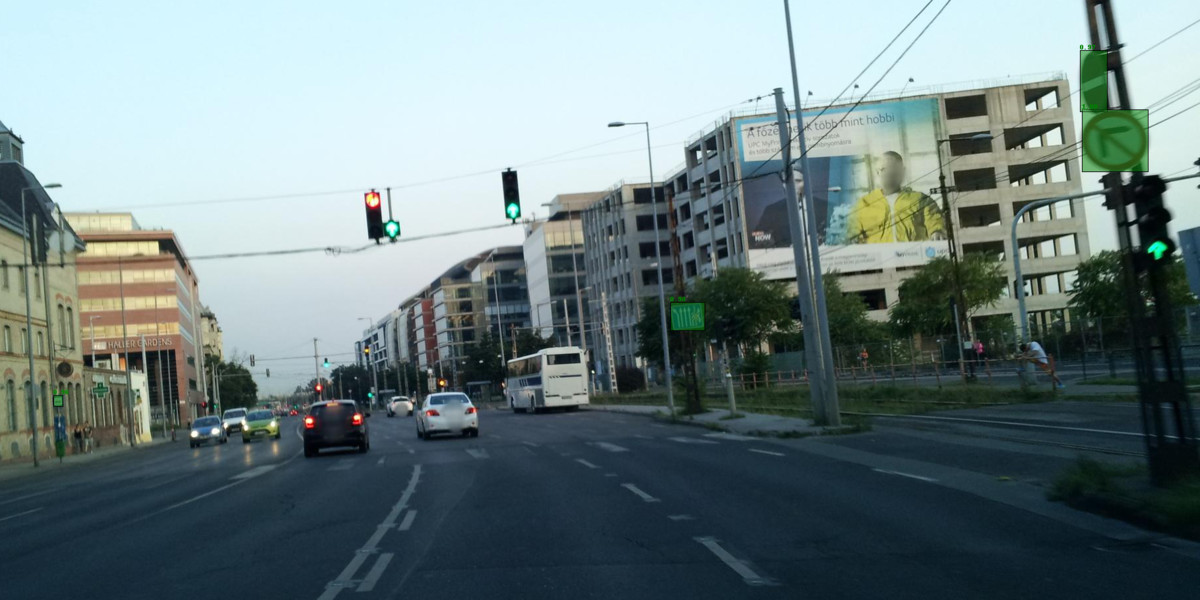}
    \end{subfigure}
    \caption{Qualitative comparisons between our binary baseline detector trained on MVD only~(left), and our baseline pre-trained on \mts\ and fine-tuned on MVD~(right). The score thresholds are chosen such that both models operate on the same level of precision.}
    \label{fig:mvd_comp}
\end{figure*}

A similar qualitative comparison for MVD is shown in \cref{fig:mvd_comp}.
Again, both models operate on the same precision level of \num{0.8}, while the model pre-trained on \mts\ obtains a higher recall of \num{0.67} compared to \num{0.61} for the model trained solely on MVD.
Besides the higher recall, the pre-trained model has less confusion with billboards and other planar objects that are similar to traffic signs.

{\small
\bibliographystyle{ieee}
\bibliography{mly}
}